\documentclass[sigconf]{acmart}

\AtBeginDocument{%
  \providecommand\BibTeX{{%
    \normalfont B\kern-0.5em{\scshape i\kern-0.25em b}\kern-0.8em\TeX}}}

\copyrightyear{2022}
\acmYear{2022}
\setcopyright{rightsretained}
\acmConference[KDD '22]{Proceedings of the 28th ACM SIGKDD Conference on Knowledge Discovery and Data Mining}{August 14--18, 2022}{Washington, DC, USA}
\acmBooktitle{Proceedings of the 28th ACM SIGKDD Conference on Knowledge Discovery and Data Mining (KDD '22), August 14--18, 2022, Washington, DC, USA}
\acmDOI{10.1145/3534678.3539110}
\acmISBN{978-1-4503-9385-0/22/08}

\usepackage{bm}
\usepackage[ruled,vlined]{algorithm2e}
\usepackage[dvipsnames]{xcolor}
\usepackage{mathtools}
\usepackage[outline]{contour}
\usepackage{lipsum}
\usepackage{diagbox}
\usepackage{tcolorbox}
\usepackage{colortbl}
\usepackage{arydshln}
\usepackage{nicefrac}
\usepackage{todonotes}
\usepackage{tabularx}
\usepackage{multirow}
\usepackage[belowskip=0pt,aboveskip=0pt]{caption}
\usepackage{tikz}
\usepackage{amsthm}
\usepackage{enumitem}
\newcommand*\circled[1]{\tikz[baseline=(char.base)]{
  \node[shape=circle,draw,inner sep=1pt] (char) {#1};}}
 
\usepackage{amsmath,amsfonts,bm}

\def\eqref#1{equation~\ref{#1}}

\def\1{\bm{1}}

\def\va{{\bm{a}}}

\def\vk{{\bm{k}}}

\def\vx{{\bm{x}}}

\def\mS{{\bm{S}}}

\DeclareMathAlphabet{\mathsfit}{\encodingdefault}{\sfdefault}{m}{sl}
\SetMathAlphabet{\mathsfit}{bold}{\encodingdefault}{\sfdefault}{bx}{n}

\def\gX{{\mathcal{X}}}

\DeclareMathOperator*{\argmin}{arg\,min}

\newcommand{\model}{\textsc{CMHE}}

\newtheoremstyle{mystyle}
{2ex}
{2ex}
{\itshape}
{0pt}
{\scshape}
{.}
{.5em}
{}

\theoremstyle{mystyle}

\newtheorem{defn}{Definition}
\newtheorem{asmp}{Assumption}

\newtheorem{remark}{Remark}

\usepackage{arydshln}

\begin{document}

\title{Counterfactual Phenotyping with Censored Time-to-Events}

\author{Chirag Nagpal$^{1}$, Mononito Goswami$^{1}$, Keith Dufendach$^{1,2}$ and Artur Dubrawski$^{1}$}%
\affiliation{%
\institution{\large $^{1}$Auton Lab, School of Computer Science, Carnegie Mellon University}}
\affiliation{%
\institution{\large $^{2}$Department of Cardiothoracic Surgery, University of Pittsburgh Medical Center}}
\email{{chiragn, mgoswami, kdufenda, awd}@andrew.cmu.edu}


\renewcommand{\shortauthors}{Nagpal, Goswami, Dufendach and Dubrawski}

\definecolor{LightCyan}{rgb}{0.88, 1, 1}
\definecolor{LightPink}{rgb}{1, 0.88, 0.88}
\definecolor{LightGray}{gray}{.8}

\begin{abstract}%
Estimation of treatment efficacy of real-world clinical interventions involves working with continuous time-to-event outcomes such as time-to-death, re-hospitalization, or a composite event that may be subject to censoring. Counterfactual reasoning in such scenarios requires decoupling the effects of confounding physiological characteristics that affect baseline survival rates from the effects of the interventions being assessed. In this paper, we present a latent variable approach to model heterogeneous treatment effects by proposing that an individual can belong to one of latent clusters with distinct response characteristics. We show that this latent structure can mediate the base survival rates and help determine the effects of an intervention. We demonstrate the ability of our approach to discover actionable phenotypes of individuals based on their treatment response on multiple large randomized clinical trials originally conducted to assess appropriate treatment strategies to reduce cardiovascular risk.
\end{abstract}

\ccsdesc[500]{Mathematics of computing~Survival analysis}
\ccsdesc[500]{Applied computing~Health informatics}
\ccsdesc[500]{Mathematics of computing~Causal networks}
\ccsdesc[500]{Computing methodologies~Latent variable models}
\keywords{Time-to-Event, Survival Analysis, Heterogeneous Treatment Effects, Phenotyping, Hazard Ratio, Subgroup Discovery}

\maketitle

\section{Introduction}

Real world studies to estimate the effect of an intervention often involve time-to-event outcomes which are typically followed up only for a fixed period of time. Such studies are commonplace in healthcare and frequently arise when evaluating the effect of a drug or medical intervention on the time to events of interest such as death, re-hospitalization, or a composite physiological outcome. 
Randomized Control Trials (RCT) aim to eliminate group imbalance through randomizing treatment and control groups. Covariates are evaluated to ensure balanced control and treatment groups so the two groups can be compared without confounding the treatment effect. Hence, in an RCT the population-level event time distributions can be directly compared to obtain estimates of average treatment efficacy. 

Indeed, popular population-level metrics for survival and time-to-event prediction involve comparing hazard ratios or summary metrics such as restricted mean time to event by building Proportional Hazard or Kaplan-Meier estimators on the treatment and control arms. 

While population-level effect estimation is important as it informs current clinical guidelines and practices, the effect of any intervention is rarely uniform across any population under observation. The advent of precision medicine aims to address these differences in treatment effect by applying individualized treatments designed based on each patient's individuals characteristics. This strategy assumes there are differences in treatment effects that may be explained by varying demographic factors, baseline physiology, or prior medical history. The crux of precision medicine thus entails careful phenotyping of individuals that may receive augmented benefit from a treatment versus those who would not benefit (or worse, suffer adverse effects). These individual patient factors can then be accounted for when framing clinical guidelines based on randomized trials. 

\contourlength{1pt}

\begin{figure*}
    \centering
    $\overbrace{\hphantom{xxxxxxxxxxxxxxxxxxxxxxxxxxxxxxxxxx}}^{\shortstack{a) \textbf{Untreated Survival Rate} ($T|\textbf{do}(A)=0$}) }\quad\overbrace{\hphantom{xxxxxxxxxxxxxxxxxxxxxxxxxxxxxxxxxxxxxxxxxxxxxxxxxxxxxxxxxxxxxxxxx}}^{\shortstack{b) \textbf{Survival Under Treatment} ($T|\textbf{do}(A)=\mathbf{1}$)} }$
    \includegraphics[height=0.25\textwidth]{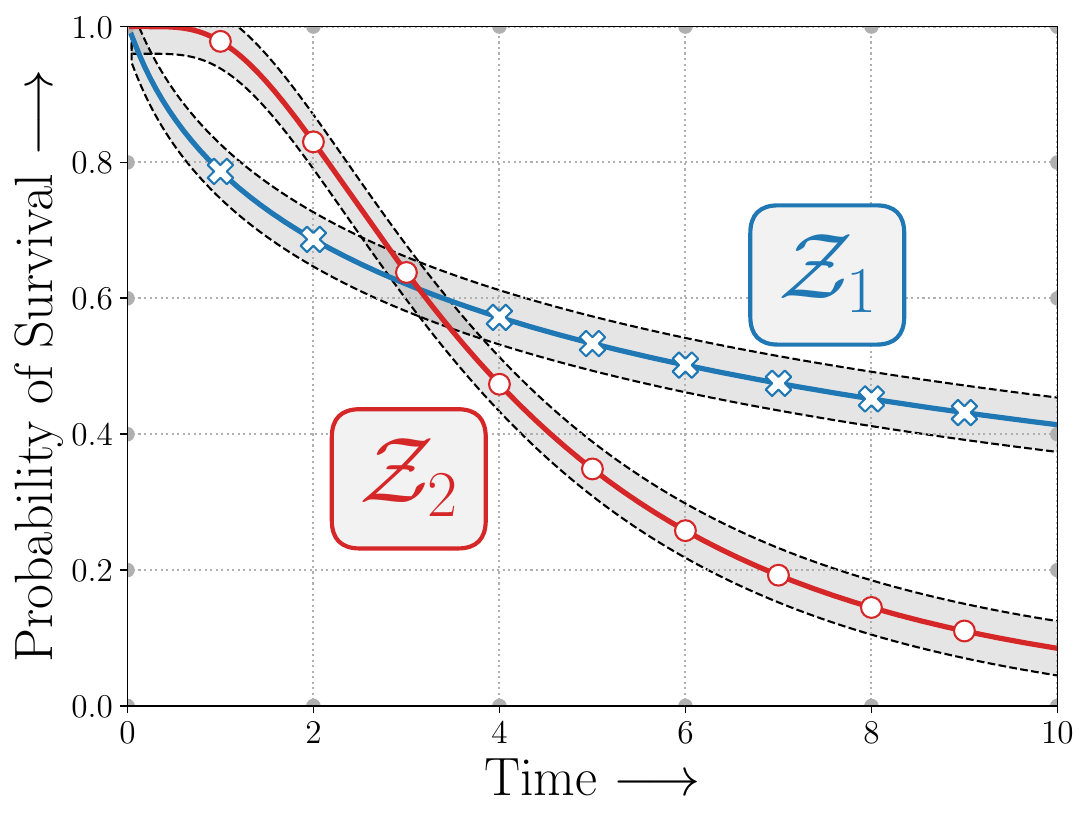}%
    \includegraphics[height=0.25\textwidth]{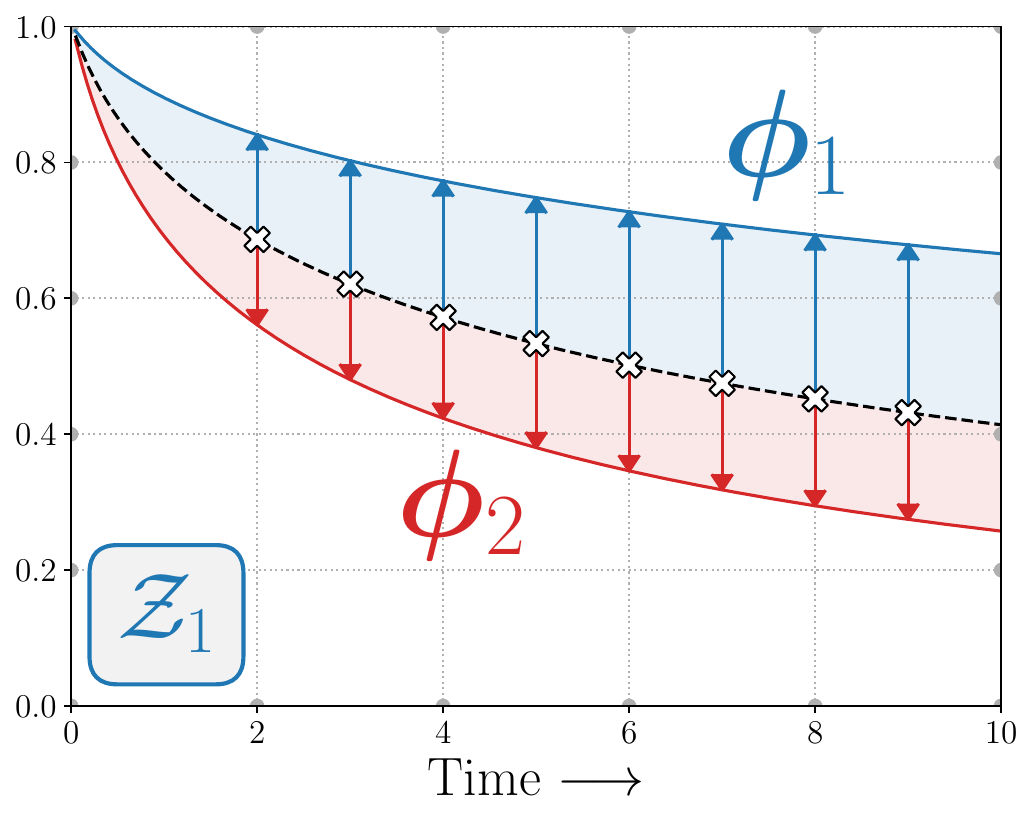}
    \includegraphics[height=0.25\textwidth]{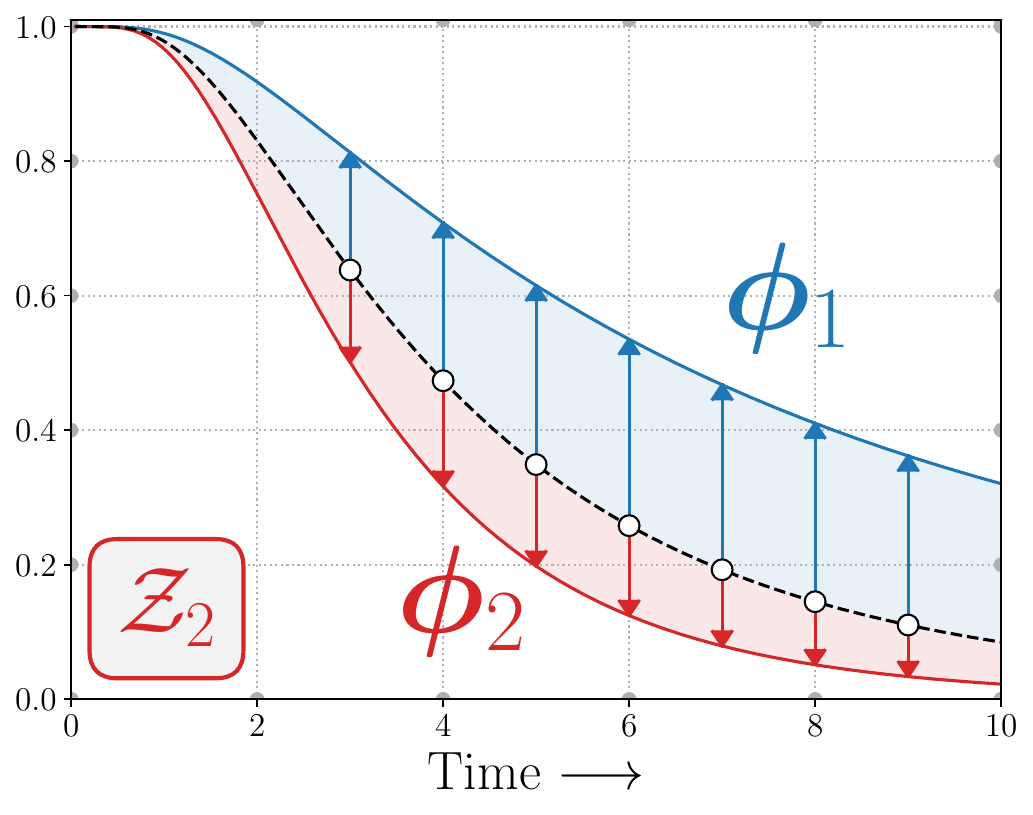}%
    \captionof{figure}{\small \textsc{Counterfactual Phenotyping}: a) In the untreated population $(T|\textbf{do}(A)=0)$ there are two subgroups ({\Large \color{white} \protect\contour{MidnightBlue}{$\mathbf{\times}$}}) and ({\Large \color{white} \protect\contour{Mahogany}{$\mathbf{\bullet}$}}) that demonstrate differential baseline survival rates represented as the latent group, $\mathcal{Z}$. The intersecting survival curves between the latent groups in $\mathcal{Z}$ suggest that the observed untreated survival rates do not obey the Proportional Hazards assumption. b) Under intervention, we observe that treatment benefit is independent of base survival group, $\mathcal{Z}$. Individuals in {\color{MidnightBlue} $\bm{\phi}_1$ benefit {\large $\bm{\uparrow}$}} while individuals in {\color{Mahogany} $\bm{\phi}_2$ are harmed {\large $\bm{\downarrow}$}}. We propose to recover such \textit{counterfactual phenotypes}, $\bm{\phi}$ that demonstrate heterogeneous effects to the intervention.}
    \label{fig:illustration}
\end{figure*}

 Consider the following scenario from an established clinical practice. 
 Landmark studies such as the ALLHAT clinical trial~\citep{davis1996rationale, cushman2002original} have established that thiazide diuretic treatment (Chlorthalidone) are not inferior to angiotensin-converting enzyme (ACE) inhibitors (Lisinopril) or calcium channel blockers (Amlodipine) for reducing cardiovascular risk in hypertensive patients. However, there are certain sub-populations, such as those with baseline chronic kidney disease, who may benefit from the renal-protective effects of an ACE inhibitor. Additionally, ACE inhibitors are not indicated as an initial treatment for hypertension in black patients, and either a thiazide diuretic or calcium channel blocker is recommended for this group~\cite{whelton20182017}.

Such scenarios demonstrate that certain risk groups or phenotypes may not benefit uniformly from a given treatment. It is therefore of immense clinical interest to recover such groups or cohorts of patients to help guide more precise interventions, which leads  to personalized medicine and improved patient safety and outcomes.
In this paper, we propose a principled approach, \textbf{Cox Mixtures with Heterogeneous Effects} to discover subgroups or cohorts of individuals that demonstrate heterogenous effects to an intervention in the presence of censored outcomes. 
The proposed method is not sensitive to strong assumptions of proportional hazards, and it can be applied in situations where such strong assumptions do not generalize uniformly across population.

Our specific contributions can be summarized as follows:
\begin{itemize}[leftmargin=1em]
    \item[$\checkmark$] We propose a deep latent variable approach
    to recover subgroups of patients that respond differentially to an intervention, in the presence of censored outcomes.
    \item[$\checkmark$] We present conditions in which the counterfactuals are identifiable using observational data under the proposed model, along with an efficient approach for learning and inference.
    \item[$\checkmark$] We demonstrate the proposed approach applied to multiple large landmark clinical trials that were originally carried out to assess the efficacy of medical interventions to reduce risk of adverse cardiovascular outcomes among hypertensive and diabetic patients, and we discover clinically actionable counterfactual phenotypes.
\end{itemize}

\textbf{Cox Mixtures with Heterogeneous Effects} has been released as part of the open-source package, \textbf{\texttt{auton-survival}} and is available at \href{https://autonlab.github.io/auton-survival/models/cmhe}{\textbf{\texttt{autonlab.github.io/auton-survival/models/cmhe}}}.

\section{Related Work}
Survival Regression, involving estimation of Time-to-Events in the presence of censored outcomes is a classic problem in statistical estimation, which has recently received attention of the machine learning research community. Arguably, the semi-parametric Cox Proportional Hazards model~\citep{cox1972regression} and its extensions involving multi layer perceptrons~\citep{faraggi1995neural, katzman2018deepsurv} remain popular, even though they are constrained by strong assumptions on the event time distributions. 

Recent research in survival analysis has focused on developing flexible estimators that can ease the restrictive assumptions of the classical Cox model. 
Non-parametric approaches have been introduced involving Random Forests \citep{ishwaran2008random} and Gaussian Processes \citep{fernandez2016gaussian, alaa2017deep}. \citep{yu2011learning} proposed to treat survival as multitask classification over discrete time horizons. Work of \citep{lee2018deephit} extends that with deep neural networks in the presence of longitudinal data \cite{lee2019dynamic}. 

Other relevant deep learning approaches include adversarial learning  \citep{chapfuwa2018adversarial} and flexible parametric mixture models \citep{nagpal2021deep, nagpal2021deepb, ranganath2016deep}. Further research into survival estimation involves modelling of competing risks, easing restrictive proportional hazard assumptions, and ensuring that models are well calibrated \citep{chapfuwa2020calibration, goldstein2020x,yadlowsky2019calibration}. The focus on estimating counterfactuals with censored data has been somewhat limited in current machine learning research literature. \cite{chapfuwa2021enabling, curth2021survite} propose using integrated probability metric
penalties \citep{shalit2017estimating} to learn overlapping representations of the treated and control populations in the presence of censored outcomes. More traditional statistical literature in counterfactual survival regression includes propensity score \citep{linden2018estimating, hassanpour2019learning, hassanpour2019counterfactual} and doubly robust estimators \cite{zhao2015doubly}.

In this paper, we focus specifically on the problem of subgroup identification and phenotyping with censored outcomes. Recent research involving phenotyping of censored time to events are restricted to factual or observational phenotypes \cite{nagpal2021dcm, chapfuwa2020sca, manduchi2021deep}. Our work, however, focuses on simultaneous discovery of latent clusters (or phenotypes) that are \textit{counterfactual}, in that they demonstrate heterogeneous effects to an intervention, while learning effective predictive models capable of capturing the uncovered heterogeneity of response functions. 

\hyperref[fig:illustration]{Figure \ref*{fig:illustration}} illustrates this `\textit{Counterfactual Phenotyping}' problem. Amongst the untreated population, the latent groups $\bm{Z}$ mediate the base survival rates. However when an intervention, $\textbf{do}(A)=\bm{1}$ is performed, the treatment effect is mediated by another latent group $\bm{\phi}$, independent of the base survival rate. Identification of such counterfactual phenotypes is of immense utility from the standpoint of clinical decision making as it can be used to administer an optimal treatment strategy to populations that are most likely to benefit. 

Machine learning techniques for recovery of subgroups with heterogeneous treatment effects has restricted focus to problems with outcomes that are either continuous or binary in nature. \cite{foster2011subgroup, vittinghoff2010estimating} propose using non-parametric estimators (Decision Trees or Random Forests) to directly regress the difference of the outcomes of the treated and control groups in a framework often called "Virtual Twins" (VT). \cite{wang2022causal} propose a sampling based approach to recover sparse rule-sets identifying subgroups with enhanced effects. More recently, \cite{nagpal2020interpretable} introduced a deep latent variable approach, Heterogeneous Effect Mixture Model. While close in spirit to our contributions, this approach 1) does not decouple effects of baseline physiology on survival from the treatment effect and 2) is incompatible with censored time-to-events and hence, cannot be applied to many real-world studies in a straightforward manner.

\begin{figure*}[!t]

\begin{minipage}{1.0\textwidth}
\section{Proposed Methodology} 
    \centering
    \includegraphics[width=1\textwidth, trim={0 16cm 0 0}, clip]{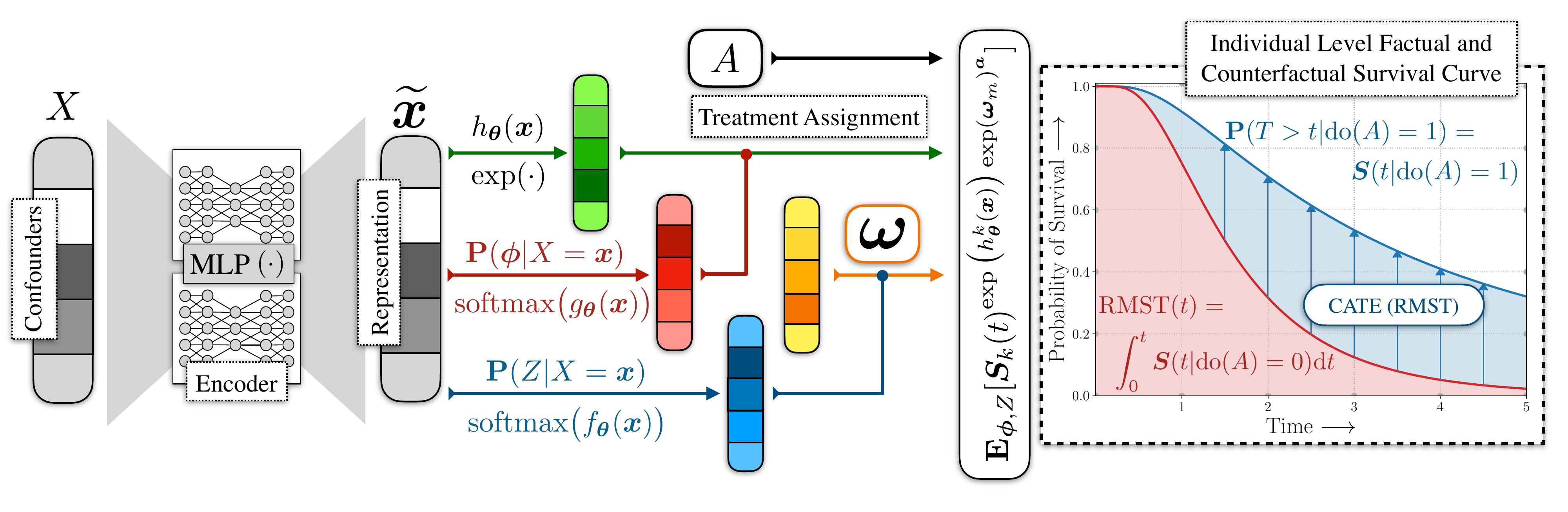}
    \captionof{figure}{\small Schematic description of the proposed \model{}. The set of features (confounders) $\bm{x}$ are passed through an encoder to obtain deep non-linear representations. These representations then describe the latent phenogroups $\mathbf{P}(Z|X=\bm{x})$ and $\mathbf{P}(\bm{\phi}|X=\bm{x})$ that determine the base survival rate and the treatment effect respectively. Finally, the individual level hazard (survival) curve under an intervention $A=\bm{a}$ is described by marginalizing over $Z$ and $\bm{\phi}$ as $\bm{S}(t|X=x, A=a) = \mathbf{E}_{(Z,\bm{\phi)}\sim \mathbf{P}(\cdot|X)}\big[ \bm{S}(t|A=\bm{a}, X, Z, \bm{\phi})\big]$.}
    \label{fig:schematic}
\end{minipage}
\end{figure*}

\subsection{Notation and Setting}
We consider a dataset of right censored observations in the form of four tuples, $\mathcal{D} = \{ (\vx_i, \delta_i, t_i, a_i ) \}_{i=1}^{N}$, where $t_i \in \mathbb{R}^+$ is either the time to event or censoring as indicated by $\delta_i\in\{0, 1\}$, $a_i\in \{0, 1 \}$ is the indicator of treatment assignment, and $\vx_i$ are individual covariates that confound the treatment assignment and the outcome.

\subsection{Cox PH Model and Cox Mixtures}
The Cox Proportional Hazards model is arguably the most popular approach to model censored survival outcomes. The Cox model involves assuming that the conditional hazard of an individual is 
\begin{align}
\bm\lambda(t|x) = \bm{\lambda}_0(t) \exp \big( h_{\bm\theta}(x) \big),
\end{align}
where $h$ is typically a linear function. Thus, under the Cox model, the full likelihood in terms of the cumulative hazard\footnote{The cumulative hazard is defined as $\bm{\Lambda}_0(t) = \int_0^{t} \bm{\lambda}_0(t)$. It can equivalently be described in terms of the base survival rate as $  \bm{\Lambda}_0(t) = - \ln \bm{S}_0(t)$.  } $\bm{\Lambda}_0$ and parameters $\bm{\theta}$ is as follows:
\begin{align}
\mathcal{L}(\bm{\theta},  \mathbf{\Lambda}_0 ) &= \prod_{i=1}^{|\mathcal{D}|} \bigg( \bm\lambda_0(t_i) \exp \big( h_{\bm \theta}(\vx_i) \big) \bigg)^{\delta_i} \mS_0(t_i) ^{\exp \big(h_{\bm \theta}(\vx_i) \big)  } 
\end{align}
However, the assumption that the hazard rates remain proportional over time is a strong one and is often violated in practice. One example of such a violation is the presence of intersecting survival curves. \cite{nagpal2021dcm} propose to relax the PH assumption by describing the data as belonging to a mixture of fixed size $K$, such that the PH assumptions hold only conditioned on the latent component assignment, $Z$. The assignment function of an individual to a latent group can then be learned jointly along with the component-specific hazard ratios. Under this model the hazard rate of an individual belonging to latent $Z=\vk$ is given as:
\begin{align}
    \nonumber \bm{\lambda}_{\bm{\theta}}(t|X=\vx, Z=\vk) &= \bm{\lambda}_k(t)\exp\big(h^{k}_{\bm\theta}(\vx)\big)\\ \text{where, } \mathbf{P}(Z&=\vk|X=\vx) \propto \exp\big(f^k_{\bm\theta}(\vx)\big) \label{eq:dcm-model} 
\end{align}

\subsection{Cox Mixtures with Heterogeneous Effects}

\begin{figure}[!t]
    \centering
    \includegraphics[width=0.35\textwidth]{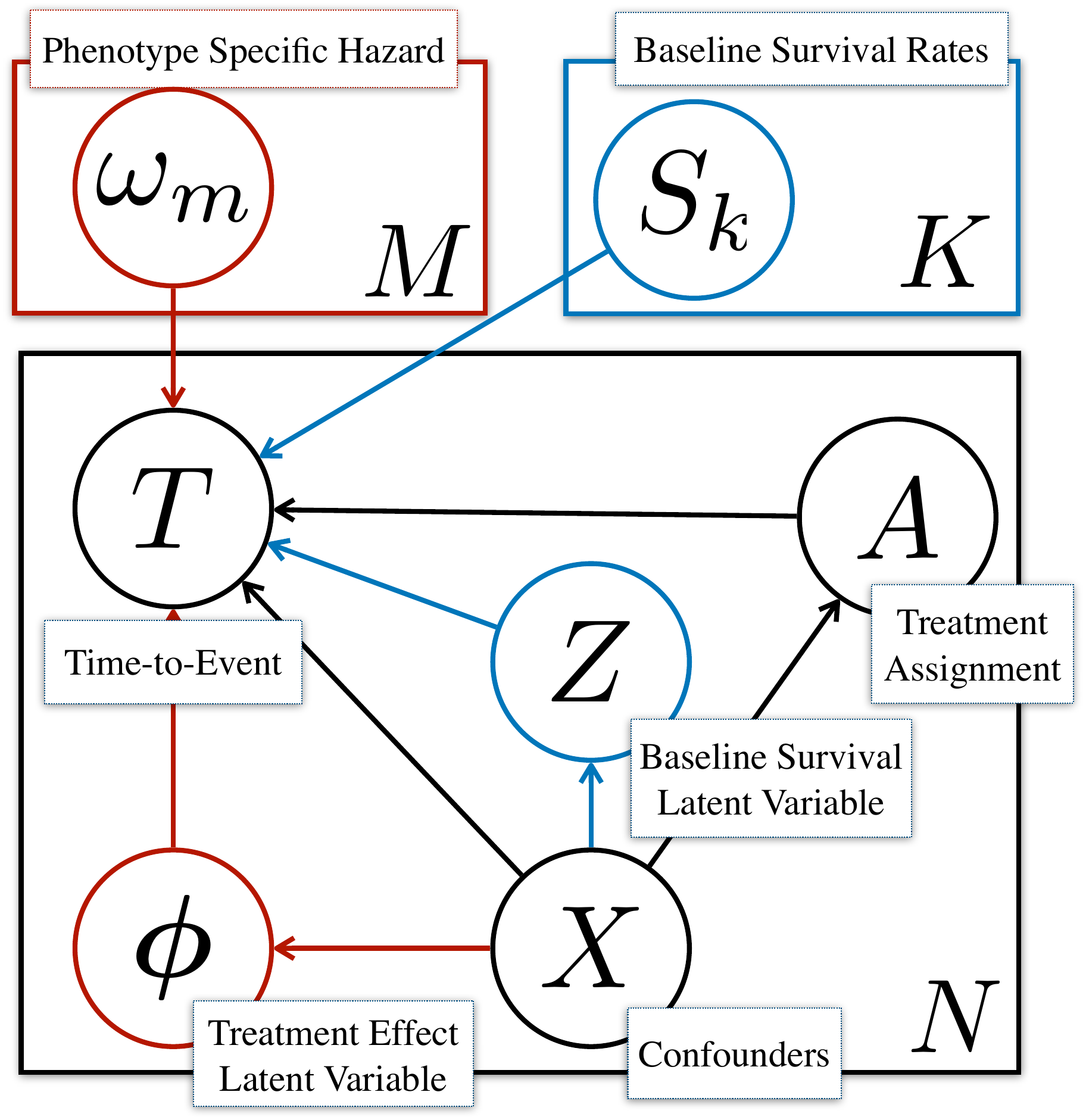}
    \captionof{figure}{\small The proposed model in Plate Notation. $X$ confounds treatment assignment $A$ and outcome $T$ (Model parameters and censoring distribution have been abstracted out).}
    \label{fig:plate-notation}
    \vspace{-1em}
\end{figure}

In this paper we show how to model counterfactual outcomes and individual level treatment effects using the proposed approach: \textbf{Cox Mixtures with  Heterogeneous Effects} (\model{}). To accomplish that, we further extend the model in \hyperref[eq:dcm-model]{ Equation \ref*{eq:dcm-model}} by introducing another latent variable $\bm{\phi}$ that determines the treatment effect for an individual with confounders, $\vx$. Thus under this new model,
\begin{gather}
\underbrace{\bm{\lambda}(t|X=\vx, {\color{MidnightBlue} Z=k},{\color{Mahogany} \bm{\phi}=m},  A=\bm{a})}_{\textrm{Conditional Hazard Rate}} = \overbrace{\bm{\lambda}_k(t)}^{\mathclap{\text{\shortstack{Base Survival Rate}}}} {\color{MidnightBlue}\underbrace{\exp\big(h^{k}_{\bm\theta}(\vx)\big)}_{\mathclap{\text{\shortstack{Effect of Confounders}}} }}{\color{Mahogany} \overbrace{\exp(\omega_m)^{\color{black}\bm{a}}}^{\mathclap{\text{\shortstack{Treatment Effect}}}}}\\
\nonumber{\color{MidnightBlue}\mathbf{P}(Z=k|X=\vx_i) \propto \exp \big( f_{\bm\theta}(\vx)\big)} \, \text{, } {\color{Mahogany}\mathbf{P}  (\bm{\phi}=m|X=\vx) \propto \exp \big( g_{\bm\theta}(\vx)\big)}.
\end{gather}
\noindent \hyperref[fig:plate-notation]{Figure \ref{fig:plate-notation}} presents \model{} in plate notation.

\noindent \begin{asmp}[Independent Censoring]
The distribution of the time-to-event $T$ and the censoring times $C$ are independent conditional on the covariates, $X$ and the treatment assignment $A$.
\label{asmp:censoring}
\end{asmp}
%
\begin{asmp}[Conditional PH] Conditional on the latent group $\mathcal{Z}$, individual time-to-event distributions obey proportional hazards.
\end{asmp}
\vspace{-1em}
\begin{asmp}[Latent Independence] The baseline survival rate group, $Z$ and the treatment effect group, $\bm{\phi}$ are independent given the confounders $X$, i.e., $\, Z \perp\bm{\phi}\, | \, X$.
\label{asmp:latent-independence}
\end{asmp}
\vspace{-1em}
\begin{asmp}[Ignorability] The treatment assignment, $A$ is independent of the potential time-to-event outcomes, $T(A)$ conditioned on the set of confounders $X$, ie. $A \perp T(A) \, | \, X$.
\label{asmp:ignorability}
\end{asmp}
Note that \hyperref[asmp:censoring]{Assumption \ref*{asmp:censoring}} is stronger compared to standard regression as it requires conditioning on both the covariates and the treatment assignment. \hyperref[asmp:ignorability]{Assumption \ref*{asmp:ignorability}} essentially states that the treatment assignment $A$ is completely characterized by the available confounders $X$. In other words, there are no exogenous factors (unobserved confounders) that may affect treatment assignments.

\begin{remark}[Identifiability] Under \hyperref[asmp:censoring]{Assumptions \ref*{asmp:censoring}} - \ref{asmp:latent-independence}, the counterfactual Time-to-Event distribution is identifiable with observables as follows,  
$\mathbf{P}(T|\textbf{do}(A)=\bm{a}, X) = \mathbf{E}_{(Z,\bm{\phi)}\sim \mathbf{P}(\cdot|X)}\big[ \mathbf{P}(T|A=\bm{a}, X, Z, \bm{\phi})\big]$.
\label{rem:identifiability}
\end{remark}
\hyperref[rem:identifiability]{Remark \ref*{rem:identifiability}} allows us to make phenogroup level counterfactual inference in terms of observables. It stems directly from the standard application of \citeauthor{pearl2009causality}'s \textbf{do}-calculus and \hyperref[asmp:latent-independence]{Assumptions \ref*{asmp:latent-independence}} and \ref{asmp:ignorability} (proof available in \hyperref[apx:identifiability-proof]{Appendix \ref*{apx:identifiability-proof}}). As a consequence of the \hyperref[asmp:censoring]{Assumptions \ref*{asmp:censoring}} - \ref{asmp:latent-independence}  under this new model, the full likelihood is:
\begin{multline}
\mathcal{L}(\bm\theta, \mathbf{\Lambda}_k) =\\
\prod_{i=1}^{|\mathcal{D}|} \underbrace{ {\color{MidnightBlue}\int_{k\in Z}} \overbrace{{\color{Mahogany}\int_{m\in \bm{\phi}}} \Big(\bm\lambda^{m}_k(t_i | \vx_i, a_i)^{\delta_i} \mS^{m}_k(t_i | \vx_i, a_i)\Big) {\color{Mahogany}\textrm{d}\mathbf{P}(\bm{\phi}|\vx_i)}}^{{\color{Mahogany}\text{Marginzalization over Treatment Effect Phenotypes, }\bm{\phi}}}{\color{MidnightBlue}\textrm{d}\mathbf{P}(Z|\vx_i)}}_{\color{MidnightBlue}\text{Marginzalization over Baseline Survival Clusters, } Z}\\
\text{where, }  \bm \lambda_k^{m}(t| \vx, a) = \bm\lambda_k(t) \exp \big( h_{\bm\theta}^k(\vx) \big) \exp \big( \bm{\omega}_{m} \big)^{\bm{a}}, \\
\mS^{m}_k(t | \vx, a) = \mS^{m}_k(t)^{\exp \big(h_{\bm\theta}^k(\vx )\big) \exp \big( \bm{\omega}_{m}\big )^{\bm{a}}}.
\label{eq:prop-model-likelihood} 
\end{multline}

 \begin{algorithm}[!t]
\small
 \caption{\textbf{ Learning for \model{}  with Stochastic EM}}
  \label{alg:algo}
\SetAlgoLined
  \SetKwInOut{Input}{Input}
  \SetKwInOut{return}{Return}
  \Input{Training set, $\mathcal{D} = \{ (\vx_{i}, t_i, a_i, \delta_i)_{i=1}^{N}  \}$; batches, $B$;} \hrulefill\\
\While{\texttt{<not converged>}}{
   \For{$b \in \{1, 2, ...,  B \} $ }{
  $\mathcal{D}_b \sim  \mathcal{D}$ $\hfill \triangleright$ Draw a minibatch from the full dataset.

 \dotfill \textbf{\textsc{{E-Step}}}

    \For{$i \in |\mathcal{D}_b|$ }{
    
    $\bm\gamma_i \sim \mathbf{P}(Z=k|\vx_i); \quad \bm \zeta_i \sim \mathbf{P}(\bm{\phi}=m|\vx_i)$ 
    
    $\hphantom{x}$     $\hfill \triangleright$ Sample soft posteriors.
    
    \vspace{0.5em}
    
    $\bm\psi_i  \sim $ \textrm{Categorical}(${\bm\gamma}_i); \quad \bm \xi_i  \sim $ \textrm{Categorical}(${\bm\zeta}_i)$

    $\hphantom{X}$ $\hfill \triangleright$ Draw hard posteriors.
    }
    \dotfill  \textbf{\textsc{M-Step}}
    
    $\bm{\theta} \gets \bm{\theta} + \eta \cdot \nabla_\theta  \widehat{Q}(\bm \theta; \mathcal{D}_b)$\\

    \hfill $\triangleright$ Update $\bm{\theta}$ with gradient of $\widehat{Q}$.

  \For{$k \in \{1, 2, ..., K \}$}{
     $ \widehat{\mathbf{\Lambda}}_k(t) \gets \sum\limits_{i:t_i<t} \bigg( {\sum \limits_ {j\in \mathcal{R}(t_i) }  \exp \big(h_{\hat{\bm{\theta}}}(\vx_j ) + {\bm{a}\cdot}\omega_{\zeta_j} \big) \bigg)^{-1} }$  \\ $ \hfill \triangleright$ \citet{breslow1972contribution}'s estimator.

    $\widehat \mS_{k}(t) \gets \exp \big( {- \widehat{\mathbf{\Lambda}}_k(t)}\big)$
    
     $\widetilde \mS_{k} \gets  \argmin\limits_s   \sum\limits_{i=1}^{n} \big(\widehat \mS_{k}(t)- s(t) \big)^2 + \lambda \int_{t_{\text{min}}}^{t_{\text{max}}} s''(t)$\\$ \hfill \triangleright $ Cubic-spline interpolation.
   }
  \dotfill 
   }
  }
\hrulefill\\
\return{learnt parameters, $\bm{\theta}$; baseline survival splines $\{\widetilde{\mS_{k}} \}_{i=1}^{K}$}
\label{alg:learning}
\end{algorithm}

\subsection{Architecture}
 
 A non-linear representation $\widetilde \vx \in \mathbb{R}^{d`}$ of the input  $\vx$ is obtained using a multilayer perceptron with parameters $\bm{\theta}$. This representation reflects the baseline survival specific log-hazard ratios, for the each of the $k$ base survival phenotypes through the function $h:\mathbb{R}^{d`}\to\mathbb{R}^k$ and the non-normalized probability of belonging to one of the $\bm{k}$ base survival clusters, ie. $\mathbf{P}(Z|X=\vx)\propto f(\vx)$ as $f:\mathbb{R}^{d`}\to\mathbb{R}^k$. Further, the non-normalized probability of assignment to counterfactual phenotype is defined as $g:\mathbb{R}^{d`}\to\mathbb{R}^m$, i.e., $\mathbf{P}(\bm{\phi}|X=\vx)\propto g(\vx)$.  \hyperref[fig:schematic]{Figure \ref*{fig:schematic}} provides a schematic diagram of the proposed architecture for \model{}.
 
\subsection{Learning}

Parameter inference in semi-parametric latent variable models such as \model{} is hard as estimation of the baseline hazard rates ($\{ \bm{\Lambda}_k\}_{k=1}^{K} $) is carried out non-parametrically. Naive application of the Expectation Maximization (EM) algorithm requires inference over all possible $(M\times K)^{|\mathcal{D}|}$ latent assignments for the entire dataset which is intractable. As described in \cite{nagpal2021deep} we propose a stochastic EM algorithm involving Monte-Carlo sampling to make inference tractable. The M-step of our proposed EM algorithm involves the following $Q(\cdot)$ function,

\begin{multline}
  \widehat{Q}(\bm \theta) =  \sum_k  \ln  \mathcal{PL}_k(\mathcal{D}_b,\bm{\psi}, \bm{\xi}; \bm\theta)+  \sum_k \sum_i^{|\mathcal{D}_b|}
 \gamma^k_i \ln  {\mathrm{softmax}\big( g_{\bm\theta}(\vx_i )\big)}\\ +   \sum_m \sum_i^{|\mathcal{D}_b|} \zeta_{i}^{m}\ln  {\mathrm{softmax}\big( h_{\bm\theta}(\vx_i )\big)},
 \label{eq:q-function}
\end{multline}
which we arrive at using the assumption that the Proportional Hazards hold within each baseline survival rate group $Z=\bm{k}$. Here $\bm\gamma$ and $\bm\zeta$ are the respective soft posterior counts of $Z$ and $\bm\phi$. $\bm{\psi}$ and $\bm{\xi}$ corresponds to the hard posterior counts sampled as $\bm{\psi} \sim \textrm{Categorical}(\bm{\gamma})$, $\bm{\xi} \sim \textrm{Categorical}(\bm{\zeta})$ and $\mathcal{PL}_k(\cdot)$ is the \textit{partial likelihood}.

 \hyperref[alg:learning]{Algorithm \ref*{alg:learning}} describes the stochastic EM learning algorithm for \model{}. The proposed stochastic EM makes inference is tractable with a complexity of $(|\mathcal{D}|\times M \times K)$. A complete discussion of our formulation along with the functional form of the $Q(\cdot)$ and $\mathcal{PL}_k(\cdot)$ functions are deferred to \hyperref[apx:learning]{Appendix \ref*{apx:learning}}

\subsection{Inference}
Following \hyperref[rem:identifiability]{Remark \ref*{rem:identifiability}}, the estimated risk of an individual with confounders $\vx$ under an intervention $\textbf{do}(A)=a$  at a time $t$ is
\begin{align}
\nonumber \widehat{\mathbf{P}}&(T>t|X=\vx, \textbf{do}(A)=\bm{a}) = \mathbf{E}_{(Z, \bm{\phi)}\sim\widehat{\mathbf{P}}(\cdot |X)}[ \widehat {\mathbf{P}}(T|X=\vx, A=\bm{a}, Z, \bm{\phi})] \\ 
\nonumber  &=  {\small \sum_k \sum_m }  {\small \textrm{softmax}}_k\big(f_{\bm\theta}(\vx)\big) \cdot {\small\textrm{softmax}}_m\big(g_{\bm\theta}(\vx)\big) \cdot \widetilde \mS^{m}_k(t)^ { \exp \big(h_{\bm\theta}(\vx) + \bm{a}\cdot\omega_m \big) }  .
\end{align}

\section{Experiments}

\begin{figure*}[!ht]

\begin{minipage}{\textwidth}
    \centering
    \begin{minipage}{0.5\textwidth}
    \centering
    \textsc{\textbf{ALLHAT}}\\
    \includegraphics[width=.75\linewidth]{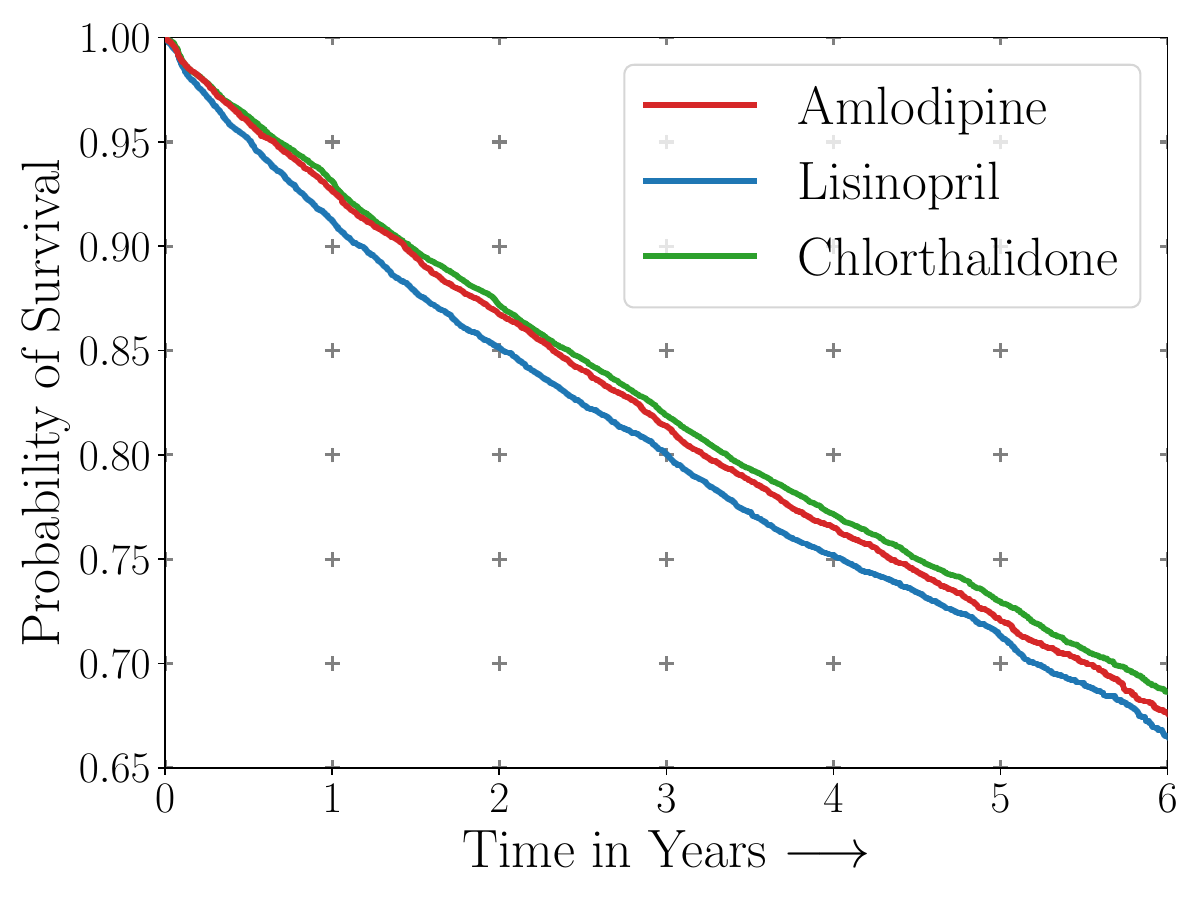}
    \end{minipage}%
    \begin{minipage}{0.5\textwidth}
    \centering
    \textsc{\textbf{ACCORD}}\\
    \includegraphics[width=.75\linewidth]{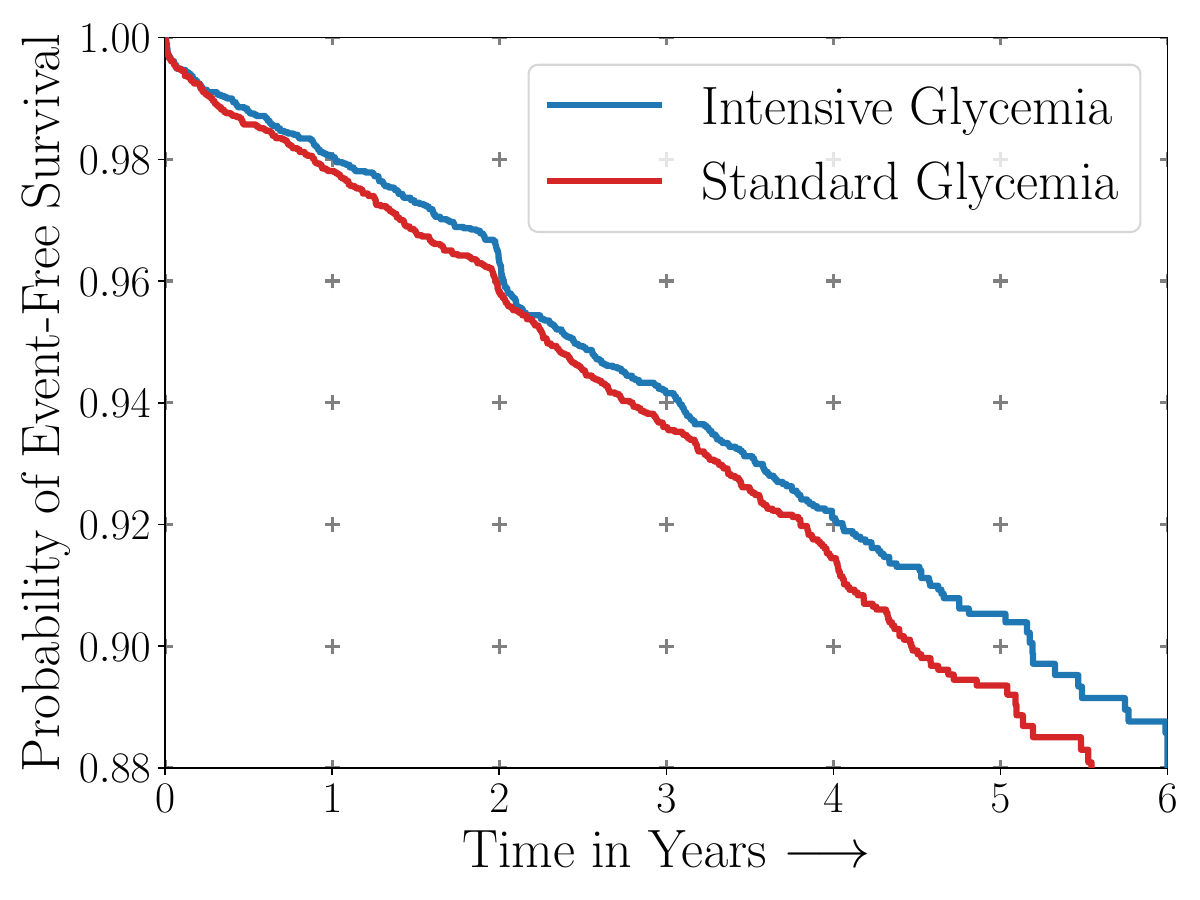}
    \end{minipage}    
\end{minipage}

\newpage

\begin{minipage}{\textwidth}
\centerline{%
\resizebox{1\textwidth}{!}{%
\begin{tabular}{|l|c|c|c|c|r|r|r|r|} \hline
\rowcolor{lightgray}\textbf{Dataset} & \textbf{Outcome} &\textbf{Treatment} & \textbf{Control} & {\small \textbf{Hazard Ratio}} &  $\textbf{{ATE}}_\textrm{RMST}(t)$ & {\small\textbf{Event Rate}} & ${N}$   \\ \hline
\textbf{ALLHAT-A} & Cardiovascular Death & Chlorothalidone & {Amlodipine/Lisinopril}&$0.94 \pm0.04$&$21.23\pm12.34$&$26.84\%$&$33,357$ \\
\textbf{ALLHAT-B}&Cardiovascular Death & Amlodipine & Lisinopril&$0.95\pm0.05$&$23.87\pm15.56$&$27.47\%$&$18,102$\\
\textbf{ACCORD}&Primary End Point & Intensive Glycemia & Standard Glycemia&$0.89\pm0.12$ &$8.28\pm14.42$& $8.50\%$ &$9,822$\\ \hline 
\end{tabular}}}
\vspace{.1em}
\captionof{figure}{ Kaplan-Meier estimates and summary statstics of the datasets used in the paper. (For ACCORD, the primary endpoint was the time to first instance of Non-Fatal Myocardial Infarction, Stroke, or Death.)}

\label{fig:data-summary}
\end{minipage}

\end{figure*}

In our experiments, we consider data from the landmark ALLHAT and ACCORD clinical trials originally conducted to determine the optimal treatment for reducing risk from cardiovascular disease. 

\subsection{Datasets}

\noindent $\blacktriangleright$ \textsc{\textbf{Antihypertensive and Lipid-Lowering Treatment to Prevent Heart Attack (\textsc{allhat})}}:
\noindent The ALLHAT clinical trial \citep{furberg2002major} was constituted to establish the appropriate intervention between chlorothalidone (a diuretic), amlodipine (calcium channel blocker) and lisinopril (angiotensin converting enzyme (ACE) inhibitor) for hypertensive patients to reduce adverse cardiovascular events. Patients were enrolled over a four year period with a mean follow up time of 4.9 years. The complete study involved 33,357 participants older than 55 years of age, all with hypertension. 15,255 ($\approx 50\%$) of participants were assigned to the Cclorthalidone arm, while 9,048 ($\approx 25\%$) were assigned to amlodipine and 9,054 ($\approx 25\%$) to lisinopril. For the purposes of this paper, we consider two separate experiments on the ALLHAT dataset.  \textbf{ALLHAT-A}: We consider all patients in the trial assigned to chlorothalidone as `Treated' and the rest as `Controls'. \textbf{ALLHAT-B}: We only consider the 18,102 assigned to lisinopril or amlodipine. Amlodipine is considered the `Treatment'. For both ALLHAT A and B we consider the time to death from cardiovascular events as the outcome of interest.\\

\begin{figure}[!t]
\centering
{\small The \textbf{SYNTHETIC} Dataset}\vspace{.15em}
\begin{minipage}{0.5\linewidth}
\centering \small $\hphantom{xxx}$ a) $\mathcal{X}$ versus $\bm{Z}$
\end{minipage}%
\begin{minipage}{0.5\linewidth}
\centering  \small $\hphantom{xxx}$ b) $\mathcal{X}$ versus $\bm{\phi}$
\end{minipage}
\includegraphics[width=0.5\textwidth, trim={0 22cm 0 0},clip]{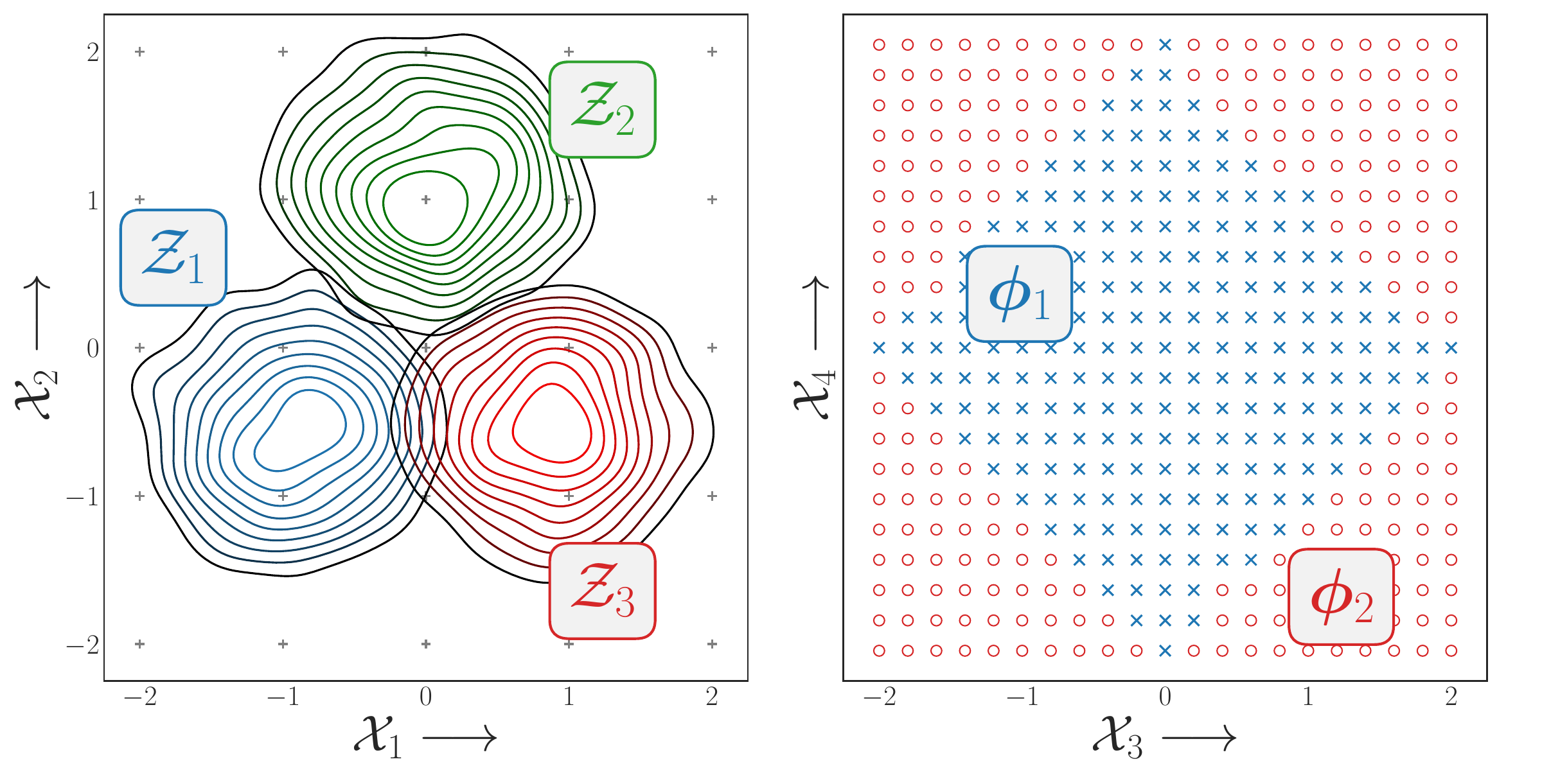}

\begin{minipage}{0.5\linewidth}\small
\centering $\hphantom{xxx}$ c) Untreated Survival
\end{minipage}%
\begin{minipage}{0.5\linewidth}
\centering \small  $\hphantom{xxxx}$ d) Survival under Treatment
\end{minipage}
\includegraphics[width=0.5\textwidth, trim={0 0 0 20.5cm},clip]{figures/synthetic/synthetic.pdf}
\caption{\small a) The distribution of the latent baseline survival groups, $Z$ in the space of features $\mathcal{X}$. b) The distribution of the latent treatment effect phenogroups $\bm{\phi}$ in the space of features $\mathcal{X}$. c) Kaplan-Meier Estimators conditioned on the latent baseline groups in the untreated population ie. $\widehat{\mathbf{P}}(T>t|Z,\textrm{do}(A)=0)$. d) Kaplan-Meier estimators of the treated population conditioned on the latent treated effect groups ie. $\widehat{\mathbf{P}}(T>t|\bm{\phi},\textrm{do}(A)=1)$.  Notice that individuals in ${\color{MidnightBlue}\bm{\phi}_1}$, {\color{MidnightBlue} benefit {\large $\bm\blacktriangle$}}  from the intervention but $\color{Mahogany}\bm{\phi}_2$ are {\color{Mahogany} harmed {\large $\bm\blacktriangledown$}}.
}
\label{fig:synthetic}%
\end{figure}

\noindent $\blacktriangleright$ \textsc{\textbf{Action to Control CVD Risk in Diabetes} (\textbf{accord}) }:
\noindent The ACCORD study \citep{ismail2010effect} involved 10,251 patients over the age of 40 with Type-2 Diabetes Mellitus with a median follow up time of 3.7 years. 5,128 ($\approx 50\%$) patients were randomized to intensive glycemic control. (\textbf{\texttt{HbA1C}\footnote{Glycated Hemoglobin (A1c)}}:$<6\%$) and the rest to standard glycemic control (\textbf{\texttt{HbA1C}}: $7-7.9\%$). In this paper we consider the patients assigned to the standard glycemic control arm as `Treated' and compare the performance in terms of time to the primary endpoint of the study, a composite endpoint including death, myocardial infarction or stroke. The \textbf{ACCORD} trial is of particular interest, as the results from it demonstrate that although intensive hyperglycemia treatment strategy for patients with diabetes reduces rate of adverse cardiovascular events, however it may increase the patient's risk for overall mortality. This is most likely due to adverse effects of the treatment itself. 

\hyperref[fig:data-summary]{Figure \ref*{fig:data-summary}} presents the Kaplan-Meier estimates of the overall population level event-free survival for the two studies along with the summary statistics including Hazard Ratio and Average Treatment Effect in Restricted Mean Survival Time. For both the \textbf{ALLHAT} and the \textbf{ACCORD} trials we consider a set of confounding features measured during the patients baseline visit at the time of randomization. This includes basic demographic information including sex and race, age at entry into the study, previous history of adverse cardiovascular events, etc. (Full list of confounders in \hyperref[sec:features]{Appendix \ref*{sec:features}}).\\

\noindent $\blacktriangleright$ \textbf{\textsc{Synthetic}}: We further benchmark the proposed model on a synthetic dataset presented in \hyperref[fig:synthetic]{Figure \ref*{fig:synthetic}}. This dataset is designed such that the latent treatment effect phenotype $\bm{\phi}$ is not linearly separable in $\bm{x}$. The time-to-event $T$ conditioned on $\bm{x}$, latent $Z$ and latent effect group $\bm{\phi}$ are generated from a Gompertz distribution. (Complete details of this design are deferred to \hyperref[apx:synthetic]{Appendix \ref*{apx:synthetic}}).

\begin{figure*}[!ht]
    \centering
    
    \textbf{Latent Phenogroups with Enhanced Treatment Effects}
    
    \includegraphics[width=0.33\textwidth]{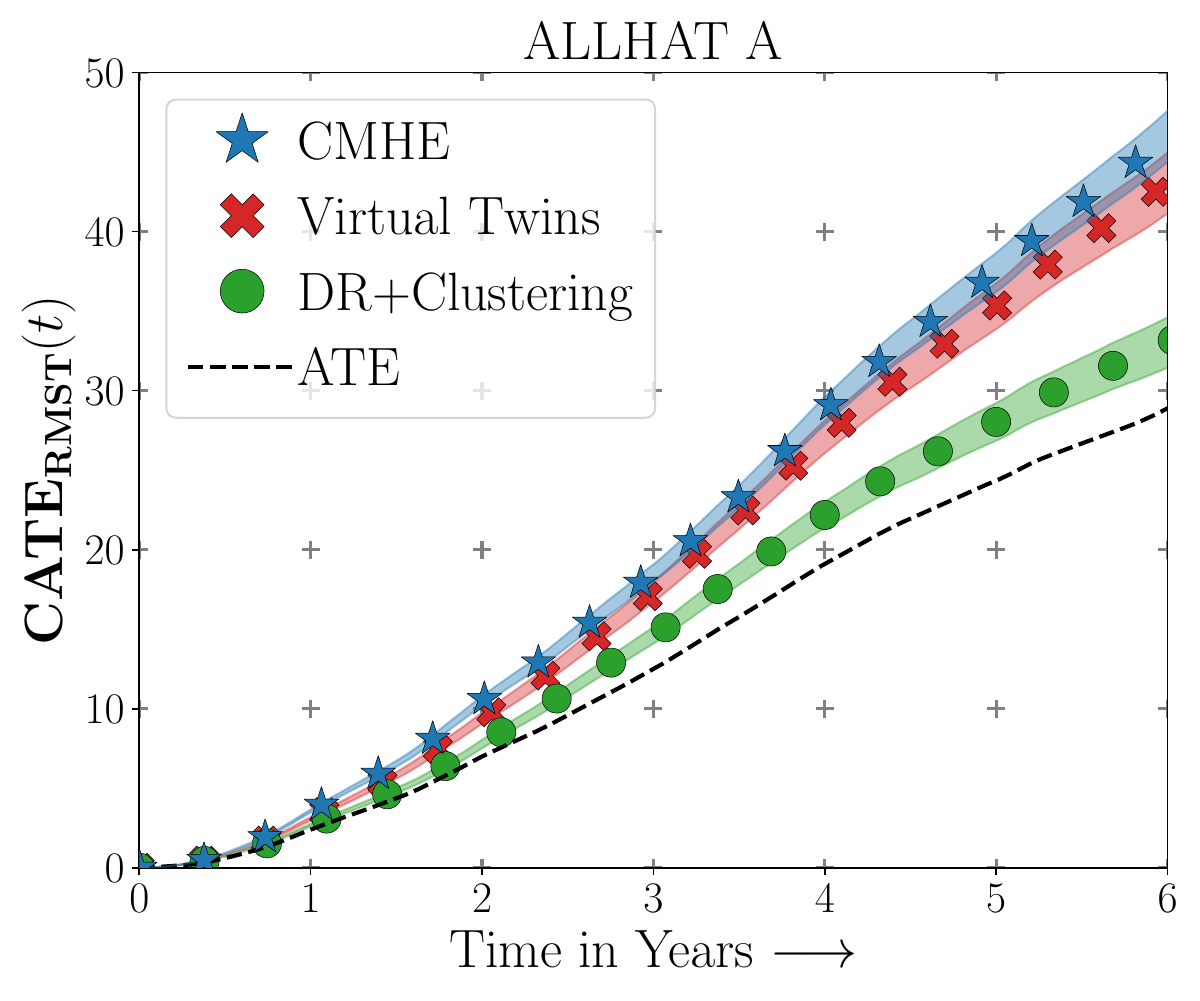}%
    \includegraphics[width=0.33\textwidth]{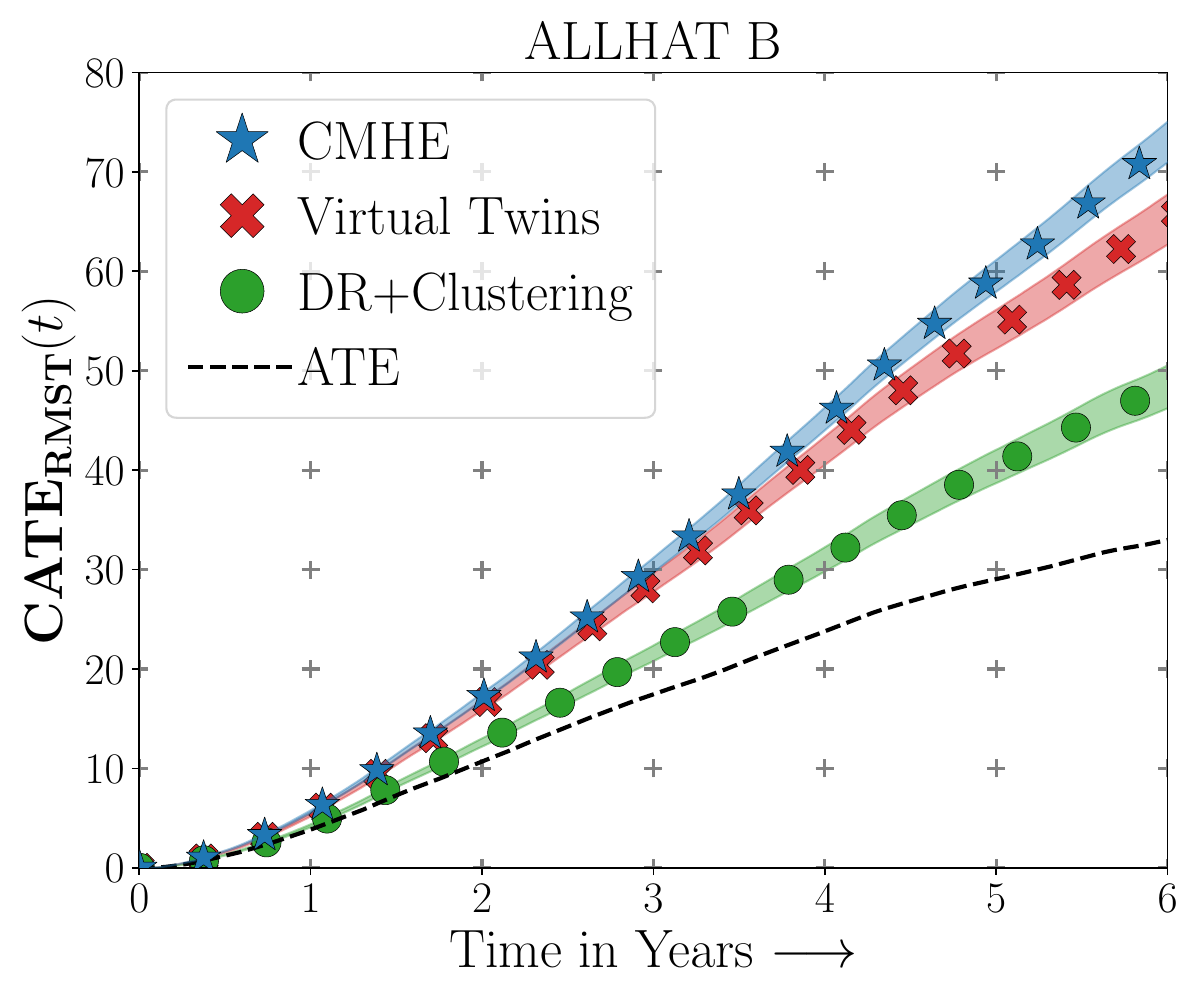}%
    \includegraphics[width=0.33\textwidth]{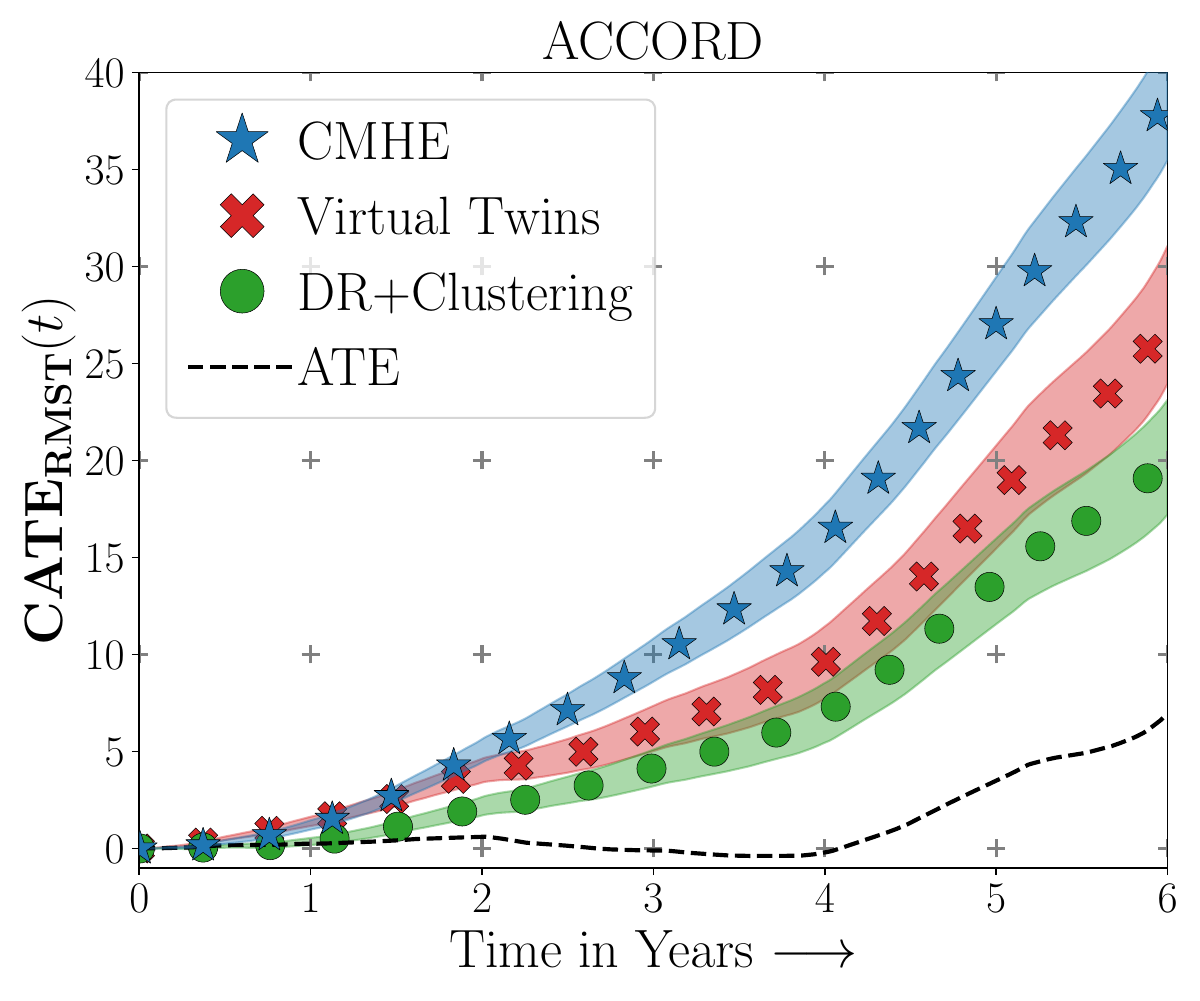}\\
    
    \textbf{Latent Phenogroups with Diminished Treatment Effects}

    \includegraphics[width=0.33\textwidth]{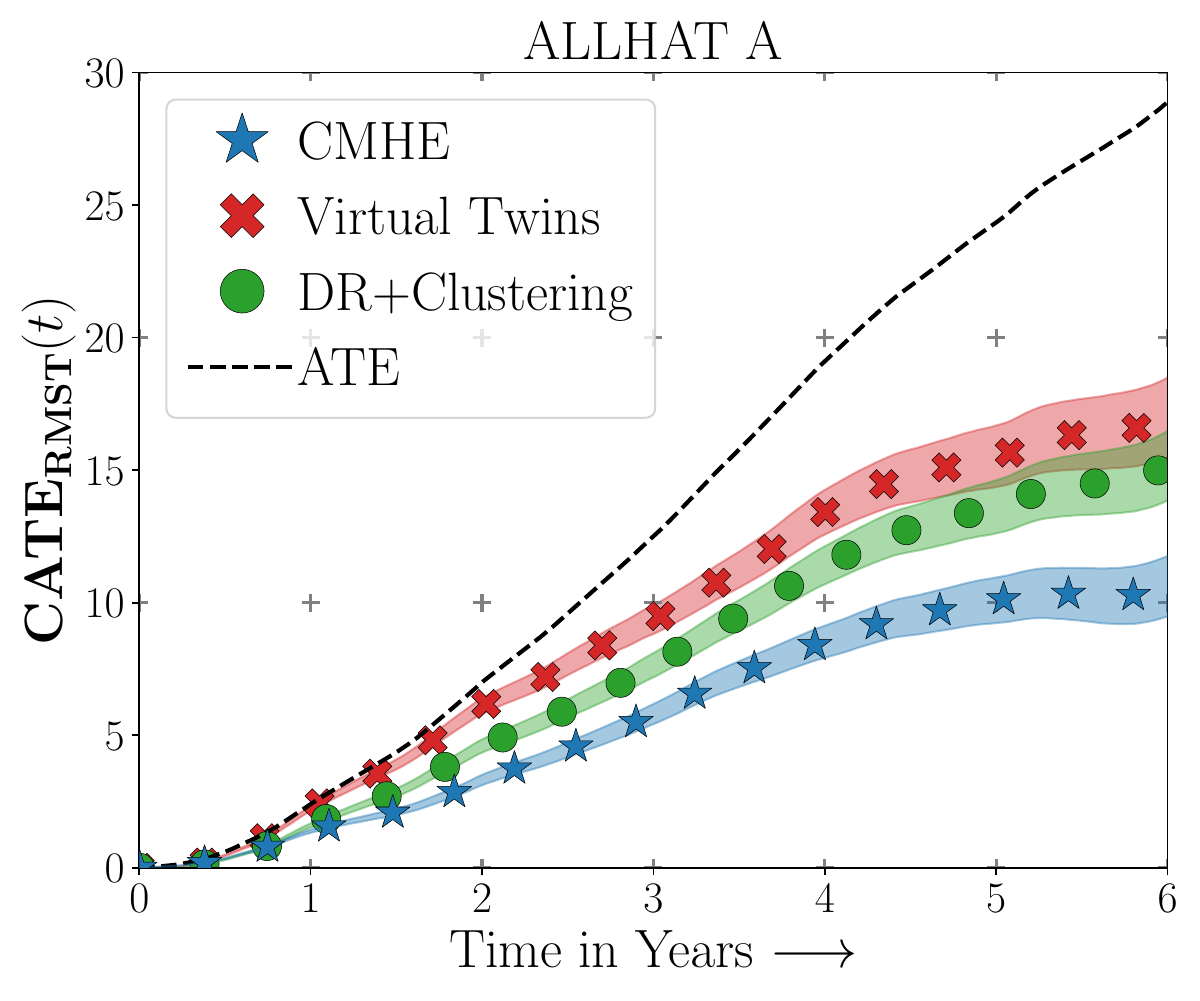}%
    \includegraphics[width=0.33\textwidth]{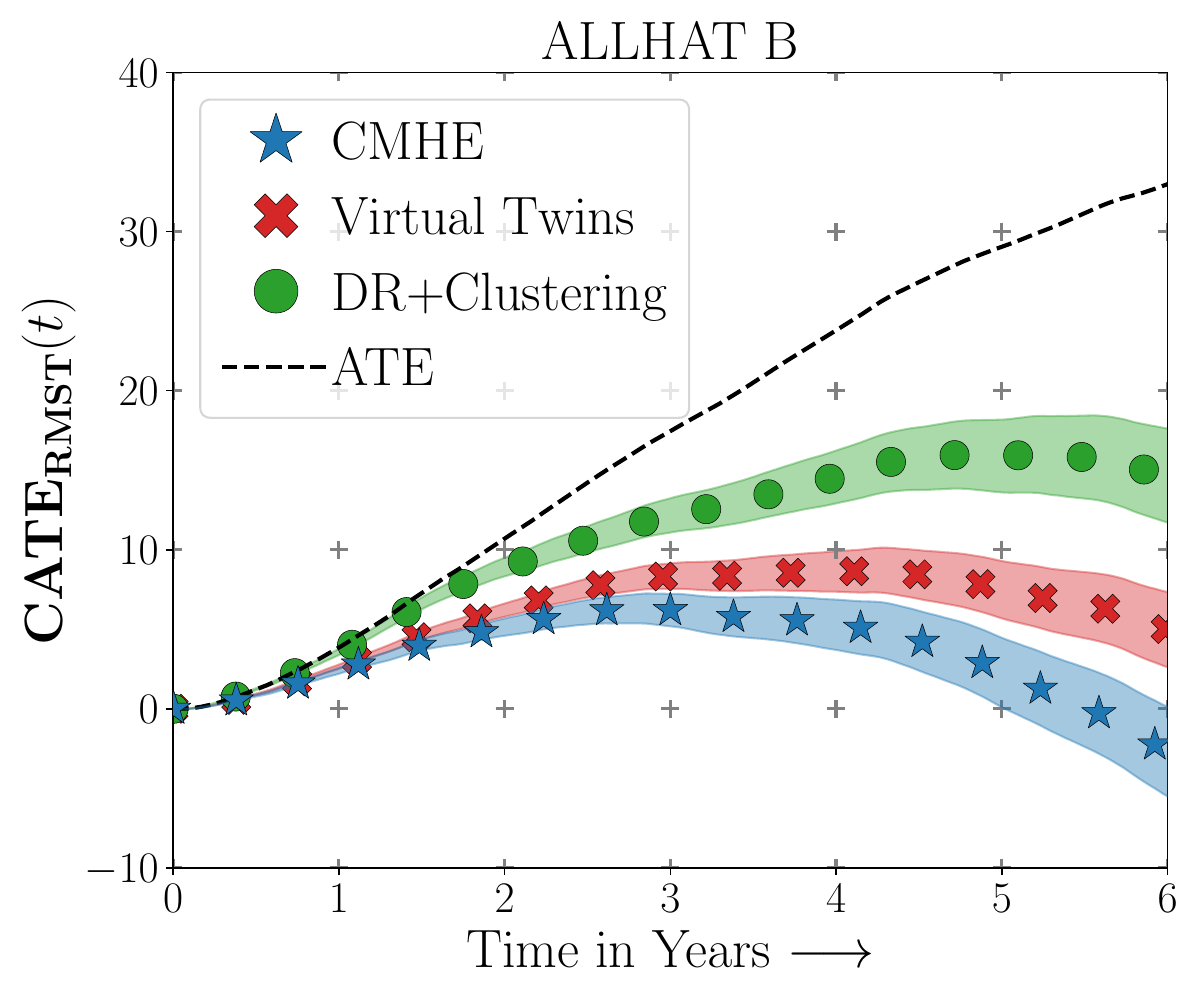}%
    \includegraphics[width=0.33\textwidth]{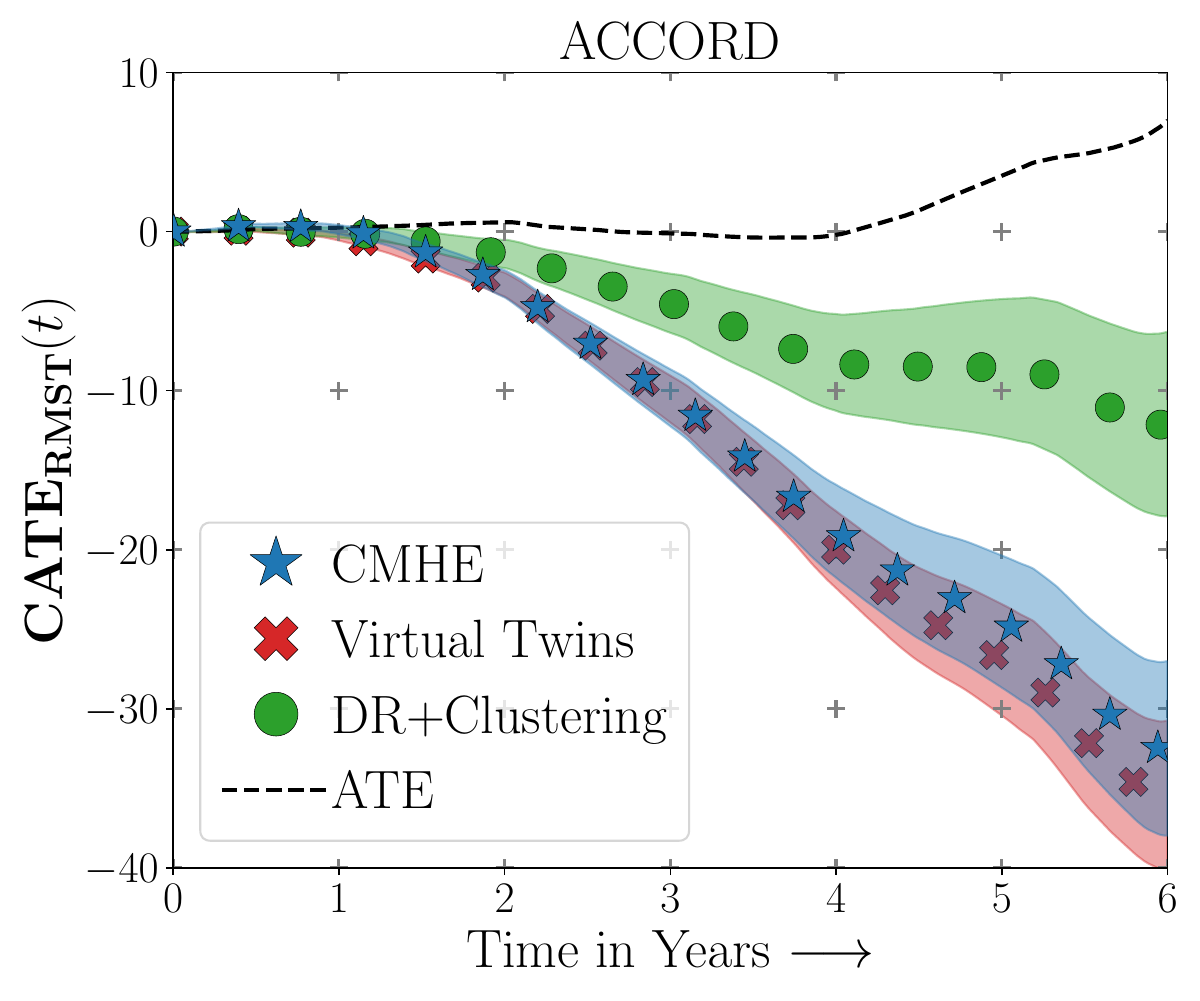}\\
    
    \caption{\small \textbf{Conditional Average Treatment Effect in Restricted Mean Survival Time} (95\% Confidence Bands) over Time for counterfactual phenotypes recovered by \model{} and Baselines in comparison to the Average Treatment Effect (ATE). For each of the datasets we identify phenogroups with enhaced (diminished) treatment effect based on RMST on the training split and report the corresponding RMST on the out-of-sample testing split. Phenogroups of different sizes are generated by varying the threshold, $\alpha>\widehat{\mathbf{P}}(\bm{\phi}|X=\vx)$ at which an individual $\vx$ is assigned to the latent phenogroup, $\mathcal{X}$. (Here we report phenogroups that are of the size 15\% of the total study population.) Notice how \model{} consistently recovers phenogroups with larger $\textbf{CATE}_\text{RMST}(t)$ (Tabulated results in \hyperref[tab:results]{Table \ref{tab:results}}).}
    \label{fig:cf-phenotypes}
\end{figure*}

\subsection{Counterfactual Phenotyping}

We evaluate the performance of \model{} in its ability to identify phenotypes of varying sizes that have a more pronounced treatment effect when compared to the baselines. For all the datasets we keep $75\%$ of the dataset as training and the rest as testing to evaluate each model's performance. For \model{} as well as the baselines we identify the phenogroup with the most enhanced and the most diminished treatment effect on the training set and report the estimated Conditional Average Treatment Effect in Restricted Mean Survival Time on the held out sample.

\begin{defn}[RMST]
The Restricted Mean Survival Time at time $\bm t$ under an intervention $\va$ for individual with confounders $\vx$ is the expected conditional Time-to-Event, \normalfont $\mathbf{E}[\textrm{min}\{T, t\}|\textrm{do}(A)=\va, X=\vx].$
\end{defn}
In the case of time-to-event outcomes, Restricted Mean Survival Time (\textbf{RMST}) is the truncated area under the survival curve,
\begin{align}
\mathbf{E}[\textrm{min}\{T, t\}|\textbf{do}(A)=\va, X=\vx] =  \int_0^{t} \mS(t|\textbf{do}(A)=\va, X=\mathbf{x})\textrm{d}t. \nonumber
\end{align}
Following from \cite{chen2001causal, royston2013restricted} we define the Conditional Average Treatment Effect in terms of the difference in Restricted Mean Survival Time under treatment and control.
\begin{defn}[$\textbf{CATE}_\text{RMST}$]
The Conditional Average Treatment Effect at time $t$ is expected difference between the treated and control Restricted Mean Survival Time conditioned on the phenogroup, $\mathcal{X}$.
\end{defn}
\vspace{-2em}
\begin{multline}
\nonumber \textbf{CATE}_\text{RMST}(\gX; t) = %
\mathop{\mathbf{E}}_{\vx \in \mathcal{X}} \big[ \mathbf{E}[\textrm{min}\{T, t\}| \textbf{do}(A)=1, X=\vx]\\ - \mathbf{E}[\textrm{min}\{T, t\}|\textbf{do}(A)=0, X=\vx]\big].
\end{multline}

This can now be estimated as:
\begin{align}
   \widehat{\textbf{CATE}}_\text{RMST}(\gX; t) = \frac{1}{n}  \sum_{x\in\mathcal{X}} \bigg[  \int_0^{t} \widehat{\mS}_1(t|\mathbf{x})\textrm{d}t - \int_0^{t} \widehat{\mS}_0(t|\mathbf{x})\textrm{d}t  \bigg].
\end{align}

Here $\gX$ is the set of all individuals in the phenotype (we control the size of the phenotype by varying the threshold, $\alpha > \widehat{\mathbf{P}}(\bm{\phi}|\vx)$ and $t$ is the time horizon at which RMST is computed. $\mS_1(\cdot)$ and  $\mS_0(\cdot)$ are the survival distributions under treatment and control.

Note that we cannot directly compare the survival curves conditioned on the recovered phenotype as within the phenotype treatment assignment is not random. In order to mitigate this problem, we fit separate Random Survival Forests (RSFs) \citep{ishwaran2008random} on the treated and control populations in the training set in a 5-fold Cross Validation to minimize the Integrated Brier Score. The fitted estimators are then employed to estimate the individual counterfactual survival curves, $\widehat{\mS}_1(t|X)$ and $\widehat{\mS}_0(t|X)$ on the test data for evaluation.\\

\noindent \textbf{\textsc{Baselines}}:
We compare the ability of \model{} against the following baseline strategies for counterfactual phenotyping:\\

\noindent $\blacktriangleright$ {\textbf{Dimensionality Reduction + Clustering}}: Involves first performing dimensionality reduction of the input confounders, $\vx$, followed by clustering. For the experiments in the paper we consider Linear-PCA and Kernel-PCA with an Radial Basis Function Kernel for dimensionality reduction, followed by K-Means and Gaussian Mixture Models (GMMs) for clustering. The number of reduced dimensions for the confounders is tuned from $\{ \textbf{\texttt{8,16}} \}$ and the number of clusters from $\{ \textbf{\texttt{2,3}} \}$. For the GMMs we enforce the learned covariance matrix of the components to be diagonal.\\

\noindent $\blacktriangleright$ {\textbf{Virtual Twins}} \citep{foster2011subgroup, vittinghoff2010estimating}: Involves first building regression estimators of the outcome conditioned on the confounders separately for the treated and control populations, followed by regressing the difference between counterfactual estimates on the confounders using a simpler model such as a Decision Tree. In the experiments, we estimate the virtual twin counterfactual survival models using a Linear Cox model and Cox model parameterized with a 2 hidden layer Multi-Layer Perceptron (MLP) with  \citep{faraggi1995neural, katzman2018deepsurv} in a 5 fold cross-validation fashion on the training dataset. For both the Linear and MLP Cox Model we tune the batch size from $\{\textbf{\texttt{128,256}} \} $ the learning rate $\{ \textbf{\texttt{2,3}} \}$ in $\{\mathbf{ 10^{-3}, 10^{-4}}\}$. For the MLP, the hidden layer was fixed to have \textbf{\texttt{Tanh}} activations with a dimensionality of \textbf{\texttt{50}}. The models were optimized using \textbf{\texttt{Adam}} \citep{kingma2014adam}. Once the counterfactual models are estimated, the difference in there estimates in terms of \textbf{RMST}@\textbf{5-Year} is modelled using a Random Forest with $\textbf{\texttt{25}}$ trees whose depth is tuned from $\textbf{\texttt{\{4,5\}}}$. The trained Random Forest is then employed to recover the counterfactual phenotypes.\\
\vspace{-0.01em}

\noindent \textbf{\textsc{Results}}: \model{} consistently recovered phenogroups that demonstrated higher CATE as compared to the the Virtual Twins and Dimensionality Reduction $+$ Clustering baselines as in \hyperref[fig:cf-phenotypes]{Figure \ref*{fig:cf-phenotypes}}. In the case of \textbf{ALLHAT A} and \textbf{B},  \model{} recovered a sub-population of 15\% of the test data that had a diminished $\textbf{CATE}_\text{RMST}$@\textbf{5-Years} of $\mathbf{10.09 \pm 0.86}$ and $\mathbf{2.28 \pm 2.18}$ Days respectively as compared to the population ATE of $\mathbf{24.30}$ and $\mathbf{28.97}$ Days (\hyperref[tab:results]{Table \ref*{tab:results}}).

\begin{figure*}[!htbp]
\centering
\begin{minipage}{0.28\linewidth}
\includegraphics[width=1\textwidth]{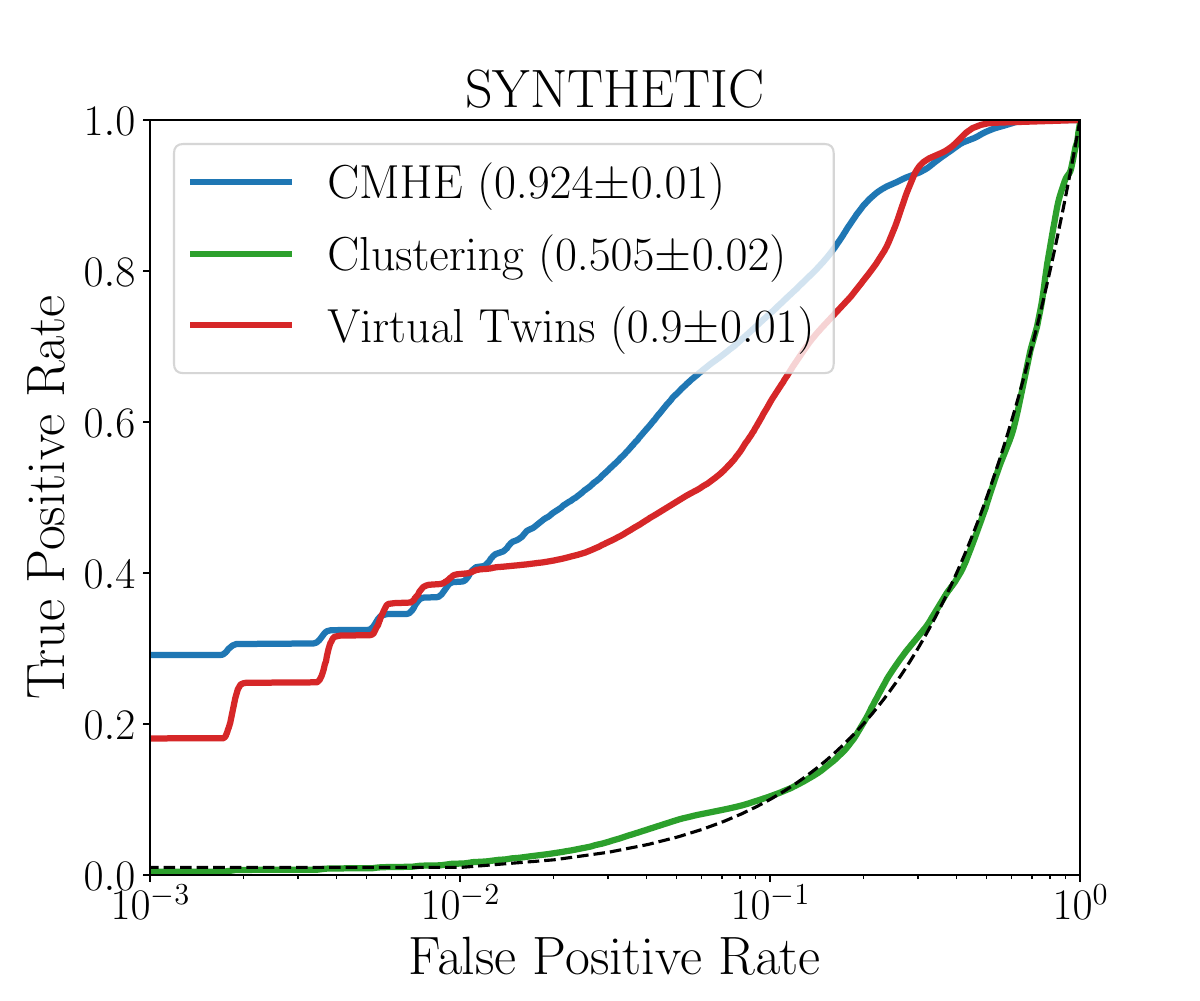}%
\captionof{figure}{\small ROC curves for recovery of latent $\bm{\phi}$ by \model{} and proposed models on the synthetic dataset.}
\label{fig:synthetic-roc}
\end{minipage}\hfill
\begin{minipage}{0.7\linewidth}
\small
\resizebox{1\linewidth}{!}{%
\begin{tabular}{|rrrr|r|r|r|} 
\hline %
\multicolumn{4}{|c|}{\multirow{2}{*}{\textbf{\large Rule explanation of \model{} Phenotype}}} &\multirow{2}{*}{{\large \textbf{Size}}}&\multicolumn{2}{c|}{\textbf{5-Year RMST}}\\ \cline{6-7}
& & & &  & $\textbf{CATE}$ & $\textbf{ATE}$  \\ \hline
 \multicolumn{7}{c}{\textbf{ALLHAT-A}} \\ \hline
\rowcolor{LightCyan}
\textbf{\texttt{Height<68.51}} & \textbf{\texttt{Dia.BP<88.5}} & \textbf{\texttt{GFR<65.7}} & \textbf{\texttt{Aspirin=+ve}}& \textbf{5.05\%}& \textbf{\textcolor{MidnightBlue}{42.87}} & \textbf{24.30} \\ 
\rowcolor{LightPink} \textbf{\texttt{Dia.BP<82.50}}  & \textbf{\texttt{Prior[T-inv.]=+ve}} & \textbf{\texttt{LLT=-ve}} & \textbf{\texttt{Race[Black]=+ve}} & \textbf{13.00\%} & \textbf{\textcolor{Mahogany}{20.32}} & \textbf{24.30} \\ \hline
\multicolumn{7}{c}{\textbf{ALLHAT-B}} \\ \hline
\rowcolor{LightCyan} \textbf{\texttt{Sys.BP>144.5}} & \textbf{\texttt{Dia.BP>87.5}} & \textbf{\texttt{GFR>55.7}} & \textbf{\texttt{Race[Black]=+ve}} &  \textbf{8.71\%} &  \textbf{\textcolor{MidnightBlue}{53.20}} & \textbf{28.97}\\ 
\rowcolor{LightPink}
\textbf{\texttt{Dia.BP<=78.5}} & \textbf{\texttt{GFR<=86.1}} & \textbf{\texttt{Race[Black]=+ve}} & \textbf{\texttt{Aspirin=-ve}} & \textbf{6.86\%} & \textbf{\textcolor{Mahogany}{7.56}} & \textbf{28.97}\\
\hline 
\multicolumn{7}{c}{\normalfont \textbf{ACCORD}} \\ \hline
\rowcolor{LightCyan} \textbf{\texttt{GFR<79.9}}  & \textbf{\texttt{UACR<42.9}} & \textbf{\texttt{FPG>194.6}} & \textbf{\texttt{Prior[CVD]=-ve}} & \textbf{5.58\%} & \textbf{\textcolor{MidnightBlue}{28.94}} & \textbf{4.79}\\ 
\rowcolor{LightPink} \textbf{\texttt{potassium<=4.8}} & \textbf{\texttt{GFR>90.2}} & \textbf{\texttt{Prior[MI]=-ve}} & \textbf{\texttt{Prior[CVD]=+ve}} & \textbf{7.56\%} & \textbf{\textcolor{Mahogany}{-23.26}} & \textbf{4.79}\\
\hline 
\end{tabular}}
\vspace{0.1em}
\captionof{table}{\small We employ a tree ensemble based rule learning algorithm (Details in \hyperref[apx:rules]{Appendix \ref{apx:rules}}) to explain the phenotypes extracted by \model{}. {\color{MidnightBlue}Enhanced} ({\color{Mahogany} Diminished}) Treatment Effect Rules in {\color{MidnightBlue} Blue} ({\color{Mahogany} Red}). We report explanations that maximize $F_1$ score on the heldout dataset. Extensive discussion on physiologic interpretation of the extracted explanations are in \hyperref[sec:interpretation]{Section \ref*{sec:interpretation}}. }
\label{tab:rules}
\end{minipage}
\end{figure*}

In the ACCORD trial, \model{} recovered a  phenogroup involving 15\% of the test set that had a much more dramatic treatment effect ($\textbf{CATE}_\text{RMST}$@\textbf{5-Years} of $\mathbf{27.02 \pm 2.43}$ Days, as compared to the population average treatment effect of $\mathbf{8.28 \pm 14.42}$ Days). These results demonstrate that \model{} can discover compelling phenogroups with substantially different treatment responses. In the case of synthetic data, we directly compare the performance of in-recovery of $\bm\phi$ by treating it as a binary classification problem. \model{} has higher discriminative performance (Area under Receiver Operating Characteristic: {$\mathbf{0.924\pm0.01}$}) versus the clustering ({$\mathbf{0.505\pm0.02}$}) and Virtual Twins ({$\mathbf{0.900\pm0.001}$}) baselines, as shown in \hyperref[fig:synthetic-roc]{Figure \ref*{fig:synthetic-roc}}. Notice that Clustering is not better than random, which is expected due to the nonlinear nature of $\bm{\phi}$ as in \hyperref[fig:synthetic]{Figure \ref*{fig:synthetic}}.\\

\noindent \textsc{\textbf{Interpretation}}:\label{sec:interpretation}
We employed a tree ensemble based rule learning \cite{friedman2008predictive} to interpret the phenotypes discovered with \model{} in terms of parsimonious conjunctions. The extracted rules are presented in \hyperref[tab:rules]{Table \ref{tab:rules}}. Additional implementation details are provided in   \hyperref[apx:rules]{Appendix \ref{apx:rules}}. The extracted rules were subjected to qualitative evaluation by an expert clinician.\\

\noindent $\blacktriangleright$ \textbf{ALLHAT-A}: Conditions associated with increased protective effect from chlorthalidone treatment include patients who are older, shorter, have decreased renal function evidenced from lower glomerular filtration rate (GFR), and exhibit less baseline cardiac disease (absence of coronary heart disease). Additionally, several conditions favor patients with lower baseline systolic or diastolic blood pressure. In clinical practice, the decision for first-line antihypertensive therapeutic agent continues to be debated \cite{whelton20182017}. However, ACE inhibitors are commonly used for treatment of patients with hypertension and cardiac disease (such as coronary heart disease). The conditions stated above are consistent with this, indicating patients with absence of coronary heart disease are more likely to benefit from diuretic therapy. Additionally, patients treated with calcium channel blockers may be prone to edema and fluid retention. Fluid retention is a risk for patients with lower baseline GFR, indicative of kidney dysfunction, thus diuretic therapy may be preferred in these patients. On the other hand, ACE inhibitors are believed to slow the progression of mild kidney disease, making them a reasonable treatment for these patients.\\

\noindent $\blacktriangleright$ \textbf{ALLHAT-B}: Patients treated with amlodipine who achieved the greatest benefit tended to have higher baseline blood pressure (systolic and diastolic), weigh less (lower weight and body mass index), have higher baseline kidney function (higher GFR), and were more likely to be of Black, non-Hispanic ethnicity. These characteristics are useful and actionable clinical parameters to guide clinicians in choosing a first antihypertensive agent. In fact, the American College of Cardiology and American Heart Association guidelines for hypertension management suggest that Black patients should be prescribed either a thiazide diuretic or calcium channel blocker for first-line agent \cite{whelton20182017}. In this case, the rules discovered with \model{} demonstrate that the calcium channel blocker amlodipine is more effective than lisinopril for Black, non-Hispanic patients. Conditions corresponding with diminished treatment effect are the inverse of those described for the group receiving the most benefit. ACE inhibitors are indicated treatment to slow the progression of renal insufficiency for those with mild renal dysfunction and hypertension, and the \model{} conditions support this, indicating amlodipine is better in patients with higher baseline kidney function, whereas lisinopril may be better for patients with kidney disease.\\

\begin{table}[!t]
    \centering
    {\begin{tabular}{|l|c|c|c|c|}
    \hline 
    \textbf{Metric}&\multicolumn{3}{c|}{\textbf{Concordance Index}} & \multirow{2}{*}{\textbf{IBS}} \\ \cline{1-4}
    \textbf{Time Horizon} & \textbf{1 Year} & \textbf{3 Year} & \textbf{5 Year}&{} \\ \hline
    \multicolumn{5}{c}{\textbf{ALLHAT-A}}\\ \hline
    Cox-Linear&0.6822&0.6716&0.6688&0.1362\\
    Cox-MLP&0.6797&0.6693&0.6677&0.1361\\ \hdashline
    \model{}-Linear&0.6830&0.6722&0.6692&0.1360\\
    \model{}-MLP&\textbf{0.6832}&\textbf{0.6734}&\textbf{0.6705}&\textbf{0.1357}\\ \hline
    \multicolumn{5}{c}{\textbf{ALLHAT-B}}\\ \hline
    Cox-Linear&0.6741&0.6641&0.6629&0.1402\\
    Cox-MLP&0.6717&0.6605&0.6602&0.1404\\ \hdashline
    \model{}-Linear&0.6753&0.6651&0.6638&0.1401\\
    \model{}-MLP&\textbf{0.6760}&\textbf{0.6655}&\textbf{0.6640}&\textbf{0.1399}\\
    \hline
    \multicolumn{5}{c}{\textbf{ACCORD}}\\ \hline
    Cox-Linear&0.6615&0.6737&0.6713&0.0591\\
    Cox-MLP&0.6564&0.6723&0.6706&0.0590\\ \hdashline
    \model{}-Linear&0.6606&0.6720&0.6697&0.0591\\
    \model{}-MLP&\textbf{0.6755}&\textbf{0.6881}&\textbf{0.6850}&\textbf{0.0587}\\ \hline
    \multicolumn{5}{c}{\textbf{SYNTHETIC}}\\ \hline
    Cox-Linear & 0.6224 & 0.6205 & 0.6158 & 0.1723\\
    Cox-MLP & 0.6623 & 0.6727 & 0.6740 & 0.1619\\ \hdashline
    \model{}-Linear & 0.6356 & 0.6365 & 0.6337 & 0.1698\\
    \model{}-MLP&\textbf{0.6676}&\textbf{0.6758}&\textbf{0.6786}&\textbf{0.1604}\\ \hline
    \end{tabular}}
    \caption{\small Time Dependent Concordance Index and Integrated Brier Score (\textbf{IBS}) for \model{} (Linear and MLP) and baseline Cox models.}
    \vspace{-1.75em}
    \label{tab:factual-results}
\end{table}

\noindent $\blacktriangleright$ \textbf{ACCORD}: \model{} revealed multiple conditions describing the phenogroup with decreased treatment effect from intense hyperglycemic control. In this group, criteria tended to favor patients with less severe renal disease (higher baseline GFR, lower baseline creatinine, lower baseline potassium) or decreased evidence of cardiovascular disease (lower diastolic blood pressure, less bradycardia), though this was not true in all cases. In contrast, the phenogroup with increased treatment effect from intense hyperglycemic control has a worse baseline kidney function (GFR < 79.9 mL/min, urine creatinine < 42.9 mg/dL), higher baseline fasted glucose levels (fasting blood glucose > 195), and lacked a documented history of cerebrovascular disease. These results suggest that patients at with decreased renal function and poor baseline glucose control, which is commonly seen in advanced diabetics, stand to gain the most from an intensive treatment regimen. Patients with a history of cerebrovascular disease (e.g., stroke), on the other hand, may be at greater risk from hypoglycemic complications related to intensive glucose control. This is consistent with clinical intuition that aggressive treatment is required for the most severe forms of a disease, whereas early stage disease may not derive as much benefit, increasing  risk for net harm due to unintended side effects.

\subsection{Factual Regression}
 For completeness, we further evaluate the performance of the proposed approach in estimating factual risk over multiple time horizons using the standard survival analysis metrics, including:
 
 \noindent \textbf{Brier Score} $\big(\textrm{BS}(t)\big)$: Defined as the Mean Squared Error (MSE) around the probabilistic prediction at a certain time horizon.
\begin{align}
\text{BS}(t) = \mathop{\mathbf{E}}_{x\sim\mathcal{D}}\big[ ||\mathbf{1}\{ T > t \} - \widehat{\mathbf{P}}(T>t|X)\big)||_{_\textbf{2}}^\textbf{2}  \big]
\end{align}
\noindent \textbf{Time Dependent Concordance Index} ($C^{\text{td}}$): A rank order statistic that computes model performance in ranking patients based on their estimated risk at a specfic time horizon.
\begin{align}
C^{td }(t) = \mathbf{P}\big( \hat{F}(t| \mathbf{x}_i) > \hat{F}(t| \mathbf{x}_j)  | \delta_i=1, T_i<T_j, T_i \leq t \big) 
\end{align}
We compute the censoring adjusted estimates of the Time Dependent Concordance Index \cite{antolini2005time, gerds2013estimating} and the Integrated Brier Score\footnote{Brier Score integrated over 1, 3 and 5 years. $\text{IBS} = \mathop{\sum}_t  \nicefrac{t}{t_\text{max}}  \cdot \text{BS}(t)$ } \cite{gerds2006consistent, graf1999assessment} to assess both discriminative performance and model calibration at multiple time horizons. For each of the datasets we perform 5-fold cross-validation over the hyperparameter grid as described in \hyperref[apx:factualgrid]{Appendix \ref*{apx:factualgrid}}, and report the performance of the hyperparameter setting with the lowest Brier Score averaged over all folds. \hyperref[tab:factual-results]{Table \ref*{tab:factual-results}} presents the discriminative performance and calibration of \model{} compared to Cox PH models in factual regression. We find that \model{} had similar or better discriminative performance than a simple Cox Model with a linear and MLP hazard functions. \model{} was also better calibrated as evidenced by overall lower Integrated Brier Score, suggesting utility for factual risk estimation.

\section{Discussion and Conclusion}

We proposed a novel deep learning approach able to discover latent phenogroups that respond differentially to an intervention in the presence of censored time-to-event outcomes. It provides a valuable adjunct to traditional statistical techniques in healthcare survival research and can aid determination of treatment efficacy. This new technique, \model{}, provides the opportunity to gain individualized patient insights by identifying subjects who would benefit from a treatment as well as those who are at highest risk for harm. This can be particularly useful in clinical practice when these personalized insights differ from population-level expectations of the efficacy of the established as well as novel treatment protocols.

\begin{acks}
This work was partially funded by the Defense Advanced Research Projects Agency under the award FA8750-17-2-0130.
\end{acks}


\bibliographystyle{ACM-Reference-Format}
\bibliography{ref_compact}
\appendix


\section{Additional Details on \model{}}

\subsection{Identifiability}
\label{apx:identifiability-proof}

Proof of \hyperref[rem:identifiability]{Remark \ref*{rem:identifiability}}. $\mathbf{P}(T|\textbf{do}(A)=\bm{a}, X) = $
\begin{multline}
\nonumber {\int_Z} \int_{\bm{\phi}}\mathbf{P}(T|\textbf{do}(A)=\bm{a}, X, Z, \bm{\phi}) \mathbf{P}(Z, \bm{\phi} |\textbf{do}(A)=\bm{a}, X)\\
= \nonumber {\int_Z} \int_{\bm{\phi}}\mathbf{P}(T|\textbf{do}(A)=\bm{a}, X, Z, \bm{\phi}) \mathbf{P}(Z|\textbf{do}(A)=\bm{a}, X) \mathbf{P}(\bm{\phi}|\textbf{do}(A)=\bm{a}, X)\\
\nonumber \text{Because, $Z \perp \phi \, | \, X$}. \text{ But, } \mathbf{P}(Z|\textbf{do}(A)=\bm{a}, X) = \mathbf{P}(Z|X) \text { and, }\\ \mathbf{P}(\bm{\phi}|\textbf{do}(A)=\bm{a}, X) = \mathbf{P}(\bm{\phi}|X)\,
(\text{From Pearl's 3$^\text{rd}$ Rule of \textbf{do}-Calculus.})\\
= \nonumber \int_{\bm{\phi}} \int_Z \mathbf{P}(T|\textbf{do}(A)=\bm{a}, X, Z, \bm{\phi}) \mathbf{P}(Z|X) \mathbf{P}(\bm{\phi}|X)\\
= \nonumber \int_{\bm{\phi}} \int_Z \mathbf{P}(T|A=\bm{a}, X, Z, \bm{\phi}) \mathbf{P}(Z|X) \mathbf{P}(\bm{\phi}|X) \, \text{(Pearl's 2$^{\text{nd}}$ Rule.)}\\
= \mathbf{E}_{(Z,\bm{\phi)}\sim \mathbf{P}(\cdot|X)}\big[ \mathbf{P}(T|A=\bm{a}, X, Z, \bm{\phi})\big]. \qquad \blacksquare
\end{multline}

\subsection{Learning}
\label{apx:learning}

We propose to maximize the likelihood in \hyperref[eq:prop-model-likelihood]{Equation \ref*{eq:prop-model-likelihood}} using a stochastic Expectation Maximization algorithm (\hyperref[alg:algo]{Algorithm \ref*{alg:algo}}).

 \textbf{E-Step:} Involves first computing the posterior counts of the joint of the latent $\bm{Z}$ and $\bm{\phi}$ as follows: 
\begin{align}
\nonumber \mathbb{E}[\mathbf{1}\{Z=k, \bm{\phi}=m\} |\{ \bm{t}, \bm{x}, \bm{a} \} ] = \mathbf{P}(Z=k, \bm{\phi}=m| \{ \bm{t}, \bm{x}, \bm{a} \})=\\
\frac{\mathbf{P}(\bm{t}|Z=k, \bm{\phi}=m, \vx, \bm{a})\cdot \mathbf{P}(Z=k, \bm{\phi}=m|\vx, \bm{a})  }{\sum_k \sum_m \mathbf{P}(\bm{t}|Z=k, \bm{\phi}=m, \vx, \bm{a})\cdot \mathbf{P}(Z=k, \bm{\phi}=m| \vx, \bm{a})}
\end{align}
Note that for the censored individuals $\mathbf{P}(t|\cdot)$ is $\mathbf{P}(T>t|\cdot)$. In practice the $\mathbf{P}(t|\cdot)$ are obtained through spline interpolation. Now the latent variable specific soft posterior counts can be computed by marginalizing out the complementary Latent Variable. 
Thus, soft posterior counts of $Z$ are $ \gamma = \sum_{\bm{\phi}\in m} \mathbf{P}(Z=k, \bm{\phi}=m| \{ \bm{t}, \bm{x}, \bm{a} \}), $ and of $\bm{\phi}$ are, $\zeta = \sum_{\bm{Z}\in k} \mathbf{P}(Z=k, \bm{\phi}=m| \{ \bm{t}, \bm{x}, \bm{a} \}) .$

 \textbf{M-Step:} as in standard EM, involves maximizing the $Q(\cdot)$ function on the data, $\mathcal{D}$ defined as: 
\begin{align}
 \nonumber Q(\bm{\theta}) = \underbrace{\sum_{i=1}^{|\mathcal{D}|} \sum_k \sum_m \gamma^{k}\cdot\zeta^{m}\cdot  \ln \mathbf{P}(\bm{t}| Z=k, \bm{\phi}=m, \vx,\bm{a})}_{\circled{\tiny A}}+\\ \underbrace{\sum_{i=1}^{|\mathcal{D}|} \sum \limits_{k}\gamma^{k} \ln   \mathbf{P}(Z=k|\vx, \bm{a}) }_{\circled{\tiny B}} + 
\underbrace{\sum_{i=1}^{|\mathcal{D}|}\sum \limits_{m} \zeta^{m} \ln \mathbf{P}(\bm{\phi}=m|\vx,\bm{a})}_{\circled{\tiny C}}.
\label{eqn:q-function}
\end{align}
Note that \circled{{\small B}} and \circled{{\small C}} can be directly optimized with a gradient based approach. However, \circled{{\small A}} is the semi-parametric Cox event rate which is hard to optimize in the presence of soft posterior weights. We instead replace the soft weights in \circled{{\small A}} with the hard posterior counts sampled as follows: $\psi \sim \textrm{Categorical}(\gamma), \, \xi \sim \textrm{Categorical}(\zeta)$.

We thus arrive at $\widehat{Q}(\cdot)$ given as,
\begin{align*}
\widehat{Q}(\bm{\theta}) =  \underbrace{\sum_{i=1}^{|\mathcal{D}|} \sum_k \sum_m \mathbf{1}_{\{\psi_i=k\}}\mathbf{1}_{\{ \xi_i=m \}} \cdot  \ln \mathbf{P}(\bm{t}| Z=k, \bm{\phi}=m,\vx,\bm{a})}_{\circled{\tiny A'}}\, + \, \circled{\Tiny B} \, + \, \circled{\Tiny C}
\end{align*}
\begin{remark}
$\widehat{Q}(\cdot)$ is an unbiased estimate of the $Q(\cdot)$ in \hyperref[eqn:q-function]{Equation \ref*{eqn:q-function}}.
\end{remark}
\noindent \textbf{Proof}. Follows immediately from the fact that $\mathbf{E}[\widehat{Q}] = Q(\cdot)$.

We can now rewrite ${\circled{{\small A'}}}$ as,
\begin{align}
\nonumber {\circled{{\small A'}}}  &= \sum_{i=1}^{|\mathcal{D}|} \sum_k  \mathbf{1}_{\{\psi_i=k\}}  \sum_m \mathbf{1}_{\{ \xi_i=m \}}\cdot  \ln \mathbf{P}(\bm{t}| Z=k, \bm{\phi}=m, \vx, \bm{a})\\
&= \sum_k \underbrace{ \sum_{i=1}^{|\mathcal{D}|} \mathbf{1}_{\{\psi_i=k\}}  \sum_m \mathbf{1}_{\{ \xi_i=m \}}\cdot  \ln \mathbf{P}(\bm{t}| Z=k, \bm{\phi}=m, \vx, \bm{a})}_{\text{Proportional Hazards, Partial Likelihood}}
\end{align}
Maximizing \circled{{\small A'}} is equal to maximizing $\mathcal{PL}_k(\cdot)$ over each $k$ where,
\begin{multline} \nonumber
\ln \mathcal{PL}_k(\mathcal{D}, \psi, \bm \xi ; \bm\theta)  =\\ \sum_{i: \delta_i =1}^{|\mathcal{D}|} \mathbf{1}_{\{ \psi_i=k\}} \bigg( h^{k}_{}(\vx_i) + {a}\omega_{\xi_i} - \ln \sum_{j \in \mathcal{R}(t_i)} \exp \big( h^{k}(\vx_j) + {a}\omega_{\xi_j}  ) \bigg).
\end{multline}
\noindent Combining $\mathcal{PL}_k(\cdot)$ with \circled{{\Tiny B}} and \circled{{\Tiny C}} we arrive at the $\widehat{Q}$ in \hyperref[eq:q-function]{Equation~\ref*{eq:q-function}}.

\section{List of Confounding features}
\label{sec:features}

\begin{table}[!h]
\small
    \centering
    \resizebox{0.4\textwidth}{!}{
    \begin{tabular}{|r|l|}
    \hline
    \textbf{Name} & \textbf{Description}\\
    \hline
    \textbf{\texttt{RACE}}& Race of Participant\\
    \textbf{\texttt{HISPANIC}}& If Participant was Hispanic\\
    \textbf{\texttt{ETHNIC}}& Ethnicity\\
    \textbf{\texttt{SEX}}& Sex of Participant\\
    \textbf{\texttt{ESTROGEN}}& Estrogen supplementation \\
    \textbf{\texttt{BLMEDS}}& Antihypertensive treatment\\
    \textbf{\texttt{MISTROKE}}&History of Stroke\\
    \textbf{\texttt{HXCABG}}&History of coronary artery bypass\\
    \textbf{\texttt{STDEPR}}& Prior ST depression/T-wave inversion\\
    \textbf{\texttt{OASCVD}}& Other atherosclerotic cardiovascular disease\\
    \textbf{\texttt{DIABETES}}&Prior history of Diabetes\\
    \textbf{\texttt{HDLLT35}} & HDL cholesterol < 35mg/dl; 2x in past 5 years\\
    \textbf{\texttt{LVHECG}}& LVH by ECG in past 2 years\\
    \textbf{\texttt{WALL25}}&LVH by ECG in past 2 years\\
    \textbf{\texttt{LCHD}}& History of CHD at baseline\\
    \textbf{\texttt{CURSMOKE}}& Current smoking status.\\
    \textbf{\texttt{ASPIRIN}}& Aspirin use\\
    \textbf{\texttt{LLT}}&Lipid-lowering trial\\
    \textbf{\texttt{RACE2}}& Race (2 groups)\\
    \textbf{\texttt{BLMEDS2}}& Antihypertensive treatment\\
    \textbf{\texttt{GEOREGN}}& Geographic Region\\
    \textbf{\texttt{AGE}}& Age upon entry \\
    \textbf{\texttt{BLWGT}}& Weight upon entry\\
    \textbf{\texttt{BLHGT}}& Height upon entry\\
    \textbf{\texttt{BLBMI}}& Body Mass Index upon entry\\
    \textbf{\texttt{BV2SBP}}&Baseline SBP\\
    \textbf{\texttt{BV2DBP}}&Baseline DBP\\
    \textbf{\texttt{EDUCAT}}& Education\\
    \textbf{\texttt{APOTAS}}& Baseline serum potassium\\
    \textbf{\texttt{BLGFR}}& Baseline est glomerular filtration rate\\
    \hline
    \end{tabular}}
    \caption{\small List of confounding variables used for experiments involving the ALLHAT dataset.}
    \label{tab:ALLHAT-feats}
\end{table}

\begin{table}[!h]
\small
    \centering
    \resizebox{0.4\textwidth}{!}{
    \begin{tabular}{|r|l|} 
    \hline
    \textbf{Name} & \textbf{Description}\\ \hline
    \textbf{\texttt{female}}&Indicator if sex is Female\\
    \textbf{\texttt{bl\_age}}&Age in years\\
    \textbf{\texttt{cvd\_hx\_bl}}&CVD History at Baseline: 0=No, 1=Yes\\
    \textbf{\texttt{raceclass}}& Race Class: White, Black, Hispanic, Other\\
    \textbf{\texttt{sbp}}& Systolic Blood Pressure (mmHg)\\
    \textbf{\texttt{dbp}}& Diastolic Blood Pressure (mmHg)\\
    \textbf{\texttt{hr}}& Heart Rate (bpm)\\
    \textbf{\texttt{x1diab}}& Diagnosis of type 2 diabetes of >3 months duration \\
    \textbf{\texttt{x2mi}}& Myocardial infarction\\
    \textbf{\texttt{x2stroke}}&Stroke\\
    \textbf{\texttt{x2angina}}&Angina/Ischemic changes (Graded Exercise/Imaging)\\
    \textbf{\texttt{cabg}}&CABG\\
    \textbf{\texttt{ptci}}&PTCI/PTCA/Atherectomy\\
    \textbf{\texttt{cvdhist}}&Participant has history of clinical CVD events\\
    \textbf{\texttt{orevasc}}&Other revascularization procedure\\
    \textbf{\texttt{x2hbac11}}&HbA1c between 7.5\% and 11.0\% inclusive\\
    \textbf{\texttt{x2hbac9}}&HbA1c between 7.5\% and 9.0\% inclusive\\
    \textbf{\texttt{x3malb}}&Micro or macro albuminuria within past 2 years\\
    \textbf{\texttt{x3lvh}}& LVH by ECG or Echocardiogram within past 2 years\\
    \textbf{\texttt{x3sten}}& Low ABI (<0.9)/>= 50\% stenosis of coronary, carotid \textbf{or},\\& lower extremity artery within past 2 years\\
    \textbf{\texttt{x4llmeds}}&On lipid lowering medication currently \textbf{or},\\ &untreated LDL-C > 130 mg/dL within past 2 years\\
    \textbf{\texttt{x4gender}}&Gender for low HDL-C within past 2 years\\
    \textbf{\texttt{x4hdlf}}&HDL-c < 50 mg/dL within past 2 years, female\\
    \textbf{\texttt{x4hdlm}}&HDL-c < 40 mg/dL within past 2 years, male\\
    \textbf{\texttt{x4bpmeds}}&Participant currently on BP medications\\
    \textbf{\texttt{x4notmed}}&Participant not on BP medication \textbf{and}, \\ & most recent BP within past 2 years\\
    \textbf{\texttt{x4smoke}}&Current cigarette smoker\\
    \textbf{\texttt{x4bmi}}&BMI > 32 kg/m2 within past 2 years\\
    \textbf{\texttt{chol}}&Total Cholesterol (mg/dL)\\
    \textbf{\texttt{trig}}&Triglycerides (mg/dL)\\
    \textbf{\texttt{vldl}}&Very low density lipoprotein (mg/dL)\\
    \textbf{\texttt{ldl}}&Low density lipoprotein (mg/dL)\\
    \textbf{\texttt{hdl}}&High density lipoprotein (mg/dL)\\
    \textbf{\texttt{fpg}}&Fasting plasma glucose (mg/dL)\\
    \textbf{\texttt{alt}}&ALT (mg/dL)\\
    \textbf{\texttt{cpk}}&CPK (mg/dL)\\
    \textbf{\texttt{potassium}}&Potassium (mmol/L)\\
    \textbf{\texttt{screat}}&Serum creatinine (mg/dL)\\
    \textbf{\texttt{gfr}}&eGFR from 4 var. MDRD eq. (ml/min/1.73 m2)\\
    \textbf{\texttt{ualb}}&Urinary albumin (mg/dL)\\
    \textbf{\texttt{ucreat}}&Urinary creatinine (mg/dL)\\
    \textbf{\texttt{uacr}}&Creatine to albumin ratio\\ \hline
    \end{tabular}}
    \caption{\small List of confounding variables used for experiments involving the ACCORD dataset.}
    \label{tab:ACCORD-feats}
\end{table}

\section{Factual Regression Experiments}
\label{apx:factualgrid}
We compare the performance of \model{} in 5 Fold CV to a Linear Cox model and a Deep Cox Model in 5 fold cross validation with a \textbf{\texttt{2 Hidden Layer MLP}} with dimensionality of \textbf{\texttt{50}} and \textbf{\texttt{Tanh}} activations. Each model was trained with \textbf{\texttt{Adam}} with learning rates tuned from $\mathbf{\{10e^{-3}, 10e^{-4} \}}$ and minibatch size of $\mathbf{\{\texttt{\textbf{128}}, \texttt{\textbf{256}}\}}$. For \model{} we tuned the number of treatment effect phenotypes from $\bm{\phi}$ from \textbf{$\mathbf{\{\texttt{2}, \texttt{3}\}}$} and the base survival rate phenogroups from $\texttt{\{\textbf{1},\textbf{2},\textbf{3}\}}$.
\section{Rule Learning}
\label{apx:rules}
We used the python package,  \textbf{\texttt{scope-rules}}\footnote{\texttt{\textbf{\href{https://github.com/scikit-learn-contrib/skope-rules}{https://github.com/scikit-learn-contrib/skope-rules}}}}to explain the learnt phenotypes with parsimonius rules. The rules were restricted to have a maximum length of $\textbf{\texttt{4}}$ and a precision of $\textbf{\texttt{0.8}}$. Explanations with the highest $F_1$ score on the train set are reported in \hyperref[tab:rules]{Table \ref{tab:rules}}.

\section{Synthetic Dataset}
\label{apx:synthetic}
We employ the python package \texttt{sklearn}\footnote{Pedregosa, Fabian, et al. "Scikit-learn: Machine learning in Python." the Journal of machine Learning research 12 (2011): 2825-2830.} to generate the confounders $\bm{x}$.

\begin{multline}
\nonumber    [\bm{x}_1, \bm{x}_2], Z \sim \texttt{sklearn.datasets.make\_blobs}(K=3)\\
\nonumber [\bm{x}_3, \bm{x}_4]  \sim \textrm{Uniform}(-2, 2)\\
\nonumber \bm{\phi}  \triangleq \mathbf{1}\{|\bm{x}_3| + |\bm{x}_4| > 2\} \\
\nonumber A \sim  \textrm{Bernoulli}(\nicefrac{1}{2})\\
\nonumber    \bm{T}^{*}| (Z=k, \bm{\phi}=m, A=\bm{a}) \sim \nonumber \textrm{Gompertz}\big(\bm{\beta}_{k}^{\top}\bm{x} +(\bm{-a}^m)\big) \\
\nonumber \delta \sim \textrm{Bernoulli}(\nicefrac{3}{4}), \quad C \sim \textrm{Uniform}(0, \bm{t}^{*})\\
\nonumber \text{if } \delta = 1: \bm{T}  = \bm{T}^* \text{else if, } \delta = 0: \bm{T}  = C.
\end{multline}

\begin{figure}[!h]
\section{Tabulated Results}
\label{apx:results}
\begin{minipage}{0.5\textwidth}
\textbf{Latent Phenogroups with Enhanced Treatment Effects}\\
\centering
\resizebox{0.75\textwidth}{!}{
\begin{tabular}{|l|c|c|c|}
\hline
\multirow{2}{*}{\textbf{Model}} & \multicolumn{3}{c|}{\textbf{CATE (RMST) in Days}}\\ \cline{2-4}
 &\textbf{1 Year} & \textbf{3 Year} & \textbf{5 Year} \\\hline 
\multicolumn{4}{c}{\textbf{ALLHAT-A}} \\ \hline
\textbf{DR-C}&2.68 $\pm$ 0.08&14.61 $\pm$ 0.52&28.03 $\pm$ 1.19\\ 
\textbf{VT}&3.08 $\pm$ 0.11&17.36 $\pm$ 0.65&35.31 $\pm$ 1.44\\ \hdashline
\textbf{\model{}}&3.58 $\pm$ 0.10&18.52 $\pm$ 0.56&37.43 $\pm$ 1.23\\
\hline
\multicolumn{4}{c}{\textbf{ALLHAT-B}} \\
\hline
\textbf{DR-C}&4.23 $\pm$ 0.15&21.57 $\pm$ 0.79&40.37 $\pm$ 1.68\\
\textbf{VT}&5.34 $\pm$ 0.17&28.68 $\pm$ 0.93&54.16 $\pm$ 1.96\\ \hdashline
\textbf{\model{}}&5.64 $\pm$ 0.13&30.44 $\pm$ 0.74&59.54 $\pm$ 1.59\\
\hline
\multicolumn{4}{c}{\textbf{ACCORD}} \\
\hline
\textbf{DR-C}&0.36 $\pm$ 0.18&4.15 $\pm$ 0.94&13.76 $\pm$ 2.21\\
\textbf{VT}&1.40 $\pm$ 0.24&6.19 $\pm$ 1.16&18.11 $\pm$ 2.65\\ \hdashline
\textbf{\model{}}&1.22 $\pm$ 0.24&9.71 $\pm$ 1.11&27.02 $\pm$ 2.43\\
\hline
\end{tabular}} \\
\textbf{Latent Phenogroups with Diminished Treatment Effects}\\
\resizebox{0.75\textwidth}{!}{
\begin{tabular}{|l|c|c|c|}
\hline
\multirow{2}{*}{\textbf{Model}} & \multicolumn{3}{c|}{\textbf{CATE (RMST) in Days}}\\ \cline{2-4}
 &\textbf{1 Year} & \textbf{3 Year} & \textbf{5 Year} \\\hline 
\multicolumn{4}{c}{\textbf{ALLHAT-A}} \\ \hline
\textbf{DR-C}&1.60 $\pm$ 0.08&7.64 $\pm$ 0.46&13.62 $\pm$ 1.0\\
\textbf{VT}&2.24 $\pm$ 0.10&9.36 $\pm$ 0.54&15.52 $\pm$ 1.16\\ \hdashline
\textbf{\model{}}&1.38 $\pm$ 0.06&5.80 $\pm$ 0.38&10.09 $\pm$ 0.86\\
\hline
\multicolumn{4}{c}{\textbf{ALLHAT-B}} \\ \hline
\textbf{DR-C}&3.57 $\pm$ 0.20&12.16 $\pm$ 1.08&15.89 $\pm$ 2.29\\
\textbf{VT}&2.64 $\pm$ 0.14&8.34 $\pm$ 0.80&7.49 $\pm$ 1.80\\ \hdashline
\textbf{\model{}}&2.39 $\pm$ 0.18&6.20 $\pm$ 1.02&2.28 $\pm$ 2.18\\
\hline
\multicolumn{4}{c}{\textbf{ACCORD}} \\ \hline
\textbf{DR-C}&-0.06 $\pm$ 0.32&-4.51 $\pm$ 1.86&-8.59 $\pm$ 4.34\\
\textbf{VT}&-0.33 $\pm$ 0.24&-10.53 $\pm$ 1.47&-26.92 $\pm$ 3.50\\ \hdashline
\textbf{\model{}}&0.12 $\pm$ 0.29&-10.45 $\pm$ 1.79&-24.52 $\pm$ 4.15\\
\hline
\end{tabular}}
\end{minipage}
\captionof{table}{\small Tabulated results of the proposed method versus baselines in counterfactual phenotyping. We report the Conditional Average Treatment Effect in Restricted Mean Survival Time over multiple time horizons.}
\label{tab:results}
\end{figure}%

\newpage

\end{document}